\documentclass[letterpaper]{article} 
\usepackage[preprint]{aaai2027}  
\usepackage[hyphens]{url}  
\usepackage{graphicx} 
\usepackage{natbib}  
\usepackage{caption} 
\usepackage{algorithm}
\usepackage{algorithmic}
\usepackage{subcaption}
\usepackage{booktabs}
\usepackage{subcaption}
\usepackage[table]{xcolor}
\definecolor{lightcyan}{RGB}{224,255,255}
\usepackage{amsmath}
\usepackage{amssymb}
\usepackage{multirow}
\usepackage{newfloat}
\usepackage{listings}
\DeclareCaptionStyle{ruled}{labelfont=normalfont,labelsep=colon,strut=off} 
\floatstyle{ruled}
\newfloat{listing}{tb}{lst}{}
\floatname{listing}{Listing}

\usepackage{booktabs}

\title{Hierarchical Data Selection via Manifold Coverage and Sparse Feature Coverage in LLM Post-training}
\author{
    Written by AAAI Press Staff\textsuperscript{\rm 1}\thanks{With help from the AAAI Publications Committee.}\\
    AAAI Style Contributions by Peter Patel Schneider,
    Sunil Issar,\\
    J. Scott Penberthy,
    George Ferguson,
    Hans Guesgen,
    Francisco Cruz\equalcontrib\corresponding,
    Marc Pujol-Gonzalez\equalcontrib\corresponding
}
\affiliations{
    \textsuperscript{\rm 1}Association for the Advancement of Artificial Intelligence\\

    1101 Pennsylvania Ave, NW Suite 300\\
    Washington, DC 20004 USA\\
    proceedings-questions@aaai.org
}

\author{
    Peng Sun\textsuperscript{\rm 1},
    Yi Yang\textsuperscript{\rm 1},
    Antong Zhang\textsuperscript{\rm 2},
    Chunxiao Li\textsuperscript{\rm 3},\\
    Yanbo Wang\textsuperscript{\rm 4},
    Dianbo Liu\textsuperscript{\rm 5},
    Xin Chen\textsuperscript{\rm 6},
    Kai Yu\textsuperscript{\rm 7},
    Lu Chen\textsuperscript{\rm 7},
    Tianfan Fu\textsuperscript{\rm 1}\corresponding
}

\affiliations{
    \textsuperscript{\rm 1}Nanjing University
    \quad
    \textsuperscript{\rm 2}Brown University
    \quad
    \textsuperscript{\rm 3}Fudan University
    \quad
    \textsuperscript{\rm 4}North University of China
    \\
    \textsuperscript{\rm 5}National University of Singapore
    \quad
    \textsuperscript{\rm 6}Suzhou Laboratory
    \quad
    \textsuperscript{\rm 7}Shanghai Jiao Tong University
}

\begin{document}

\maketitle

\begin{abstract}
As supervised fine-tuning data continues to scale, selecting high-value subsets from large candidate pools is crucial for reducing training cost and improving model performance. Existing methods often measure diversity directly in the original embedding space, where geometric metrics entangle dominant semantic directions, fine-grained supervision differences, and local noise. We address this limitation by formulating data selection as a coarse-to-fine hierarchical coverage problem and propose MASS. MASS learns low-dimensional principal manifold coordinates with a dense autoencoder for coarse semantic grouping, and then performs quality-aware sparse feature coverage within each group using a TopK sparse autoencoder. Experiments on Vision Flan and LLaVA-CoT show that MASS consistently outperforms strong data selection baselines across multiple budgets, and in several settings matches or surpasses full data training with only a small subset of data.
\end{abstract}


\section{Introduction}

As the scale of supervised fine-tuning data for large models continues to grow, selecting high-value subsets from large candidate pools has become an important problem for reducing training cost and improving model performance~\cite{ivison2025large,liu2025less}. Existing data selection methods often use embeddings to characterize data distributions, and maintain sample diversity and coverage through similarity measurement and clustering~\cite{deb2025fishersft,bi2025prism}. Some methods further incorporate quality or importance signals to improve the effectiveness of the selected subset~\cite{lee2024concept,yan2025coido}. Although these methods are effective in practice, most of them still rely on geometric relations in the original embedding space, implicitly assuming that neighborhood relations and aggregation patterns in this space are aligned with the semantic coverage and supervision differences required for data selection.

However, we find that this assumption is insufficient. For complex instruction data, geometric relations in the original embedding space do not only reflect dominant semantic directions, but are also affected by fine-grained attributes within these directions, response format variations, and local noise~\cite{chen2025mig}. Our embedding geometry analysis further shows that the data distribution has a high degree of global variation and is difficult to summarize with a small number of linear directions. In contrast, local neighborhoods exhibit substantially lower effective dimensionality, and their neighborhood structure is more consistent with a nonlinear manifold assumption, as shown in Appendix~\ref{app:embedding_geometry}. Therefore, performing flat diversity measurement directly in the original embedding space may mix global semantic structure, local fine-grained differences, and task irrelevant perturbations into a single selection criterion.
Based on this observation, we argue that data selection should be formulated as a coarse-to-fine hierarchical coverage problem: the selected subset should first preserve coverage over major semantic manifold regions, and then further cover fine grained supervision features within each region. To this end, we propose MASS, a manifold aware sparse selection method. MASS first uses a dense autoencoder to learn stable low-dimensional principal manifold coordinates, and performs coarse grained grouping in this space. It then uses a TopK sparse autoencoder to extract sparse features, and performs fine grained coverage selection within each coarse grained group. In addition, MASS incorporates external quality scores to avoid selecting low quality samples merely for increasing diversity.

We evaluate MASS on Vision-Flan and LLaVA-CoT, covering both general instruction and reasoning tasks. Experimental results show that MASS consistently outperforms existing data selection baselines under multiple data budgets, and in some settings even surpasses full data training performance using only a small subset of data. Further ablation studies show that the principal manifold coordinates learned by the DAE provide more reliable coarse grained grouping, the sparse features learned by the SAE improve fine-grained coverage within each group, and quality signals further enhance the effectiveness of the selected samples.

Our main contributions are summarized as follows:
\begin{itemize}
    \item We reformulate supervised fine-tuning data selection as a hierarchical coverage problem that combines coarse-grained principal manifold coverage with fine-grained supervision feature coverage.
    \item We propose MASS, which constructs principal manifold groups with a DAE, performs within-group sparse feature coverage with a TopK SAE, and incorporates quality signals for sample selection.
    \item We validate MASS across multiple datasets, data budgets, embedding sources, and target models, demonstrating its effectiveness and robustness.
\end{itemize}

\section{Related Work}
\paragraph{Data Selection.}
Data selection aims to identify a high-value subset from large-scale training data, reducing training cost while preserving or even improving model performance. Existing methods usually select data from the perspective of sample importance or coverage diversity: the former includes methods based on training dynamics, such as EL2N~\cite{paul2021deep}, LESS~\cite{xia2024lessselectinginfluentialdata}, and OPUS~\cite{wang2026opus}, as well as methods based on a model's own behavior or performance, such as ScalSelect~\cite{wu2026scalselect} and CVS~\cite{sun2026does}; the latter includes representation similarity based methods, such as SemDeDup~\cite{abbas2023semdedup} and PRISM~\cite{bi2025prism}, and diversity consensus based methods, such as ICONS~\cite{wu2024icons} and INSTAG~\cite{lu2023instag}. In addition, methods such as COINCIDE~\cite{lee2024concept} and XMAS~\cite{naharas2025data} further combine sample importance with data diversity, and have shown more stable selection performance in practice. Although effective, these methods mostly rely on global scores or distribution level similarity measures, whereas MASS formulates data selection as a coarse-to-fine hierarchical coverage problem that preserves principal manifold coverage, models local fine-grained supervision feature coverage, and uses sample quality as an auxiliary signal.

\paragraph{Autoencoders and Manifold Representation.}
Autoencoders compress high-dimensional inputs into a low-dimensional latent space through an architecture with an encoder and a decoder, and learn compact representations via a reconstruction objective~\cite{baldi1989neural,hinton2006reducing}. Previous studies show that low-dimensional latent representations can preserve the principal directions of variation in data, and thus have been widely used for nonlinear dimensionality reduction, representation learning, and manifold structure modeling~\cite{kramer1991nonlinear,hinton2006reducing,bengio2013representation}. In MASS, we adopt a dense autoencoder (DAE) to compress condition representations and learn low-dimensional principal manifold coordinates, which are used to construct stable coarse-grained semantic groups. Compared with clustering directly in the original embedding space, the bottleneck structure of DAE helps suppress redundant dimensions and local perturbations, thereby better characterizing the dominant structure of the data distribution.

\paragraph{Sparse Autoencoders.}
Sparse autoencoders originate from sparse coding and dictionary learning, aiming to reconstruct input representations with a small number of activated features and thereby obtain more selective and interpretable latent features~\cite{olshausen1997sparse,aharon2006k}. Recently, SAEs have been widely used for mechanistic interpretability of language models, where they decompose internal model activations into more semantically coherent features~\cite{huben2024sparse,bricken2023towards}. Subsequent works further propose TopK SAE~\cite{makhzani2013k}, Gated SAE~\cite{rajamanoharan2024improving}, and JumpReLU SAE~\cite{rajamanoharan2024jumping} to improve the tradeoff among sparsity control, dead features, and reconstruction quality. Unlike prior works that mainly focus on interpreting internal model activations, MASS applies TopK SAE to joint input response supervision representations and uses sparse feature coverage gain for data selection, thereby improving the diversity of fine-grained supervision signals within each principal manifold region.

\section{Method}

\subsection{Problem Formulation}
Given a dataset $D$, data selection seeks a budgeted subset $D'\subseteq D$ with $|D'|=P$, such that training on $D'$ approaches or even surpasses the performance of training on $D$. 
Let $f(\cdot)$ denote the resulting model performance. 
We formulate the objective as
\begin{equation}
    \max_{D'\subseteq D} f(D')
    \quad
    \mathrm{s.t.}
    \quad
    |D'| = P .
\end{equation}

\subsection{MASS: Manifold Aware Sparse Selection}
\label{subsec:detail_mass}
We propose MASS, a manifold aware sparse selection method for SFT data selection. MASS uses a dense autoencoder (DAE) to learn low-dimensional principal manifold coordinates for coarse semantic grouping, and a TopK sparse autoencoder (SAE) to capture fine grained sparse supervision features. It then performs greedy selection within each group by combining sparse feature coverage with external quality scores, preserving manifold coverage while improving fine grained diversity and sample quality. The overall pipeline is shown in Figure~\ref{fig:mass_pipeline}, with implementation details provided in Appendix~\ref{app:data_dpo_implementation}.

\paragraph{Dual View Encoding.}
Given a candidate sample $z_i$, we represent it as an input and response pair:
\begin{equation}
z_i = (q_i, a_i),
\end{equation}
where $q_i$ denotes the input, which may contain textual instructions, visual information, or other contextual content, and $a_i$ denotes the response. For each sample, we construct two types of embeddings. The first one is the condition embedding:
\begin{equation}
e_i^{c} = E_{\mathrm{emb}}(q_i),
\end{equation}
which encodes only the semantic information contained in the input, and is used for subsequent DAE manifold coordinate learning and coarse grained grouping. The second one is the supervision embedding:
\begin{equation}
e_i^{s} = E_{\mathrm{emb}}(q_i, a_i),
\end{equation}
which encodes the complete supervision signal jointly formed by the input and the response, and is used for subsequent SAE sparse feature learning and fine grained coverage within each group. Both embeddings are extracted with a frozen embedding model, then independently centered and $L_2$-normalized. In the following sections, $e_i^c$ and $e_i^s$ denote the preprocessed embeddings.

\begin{figure*}[t]
  \centering
  \includegraphics[width=1.0\linewidth]{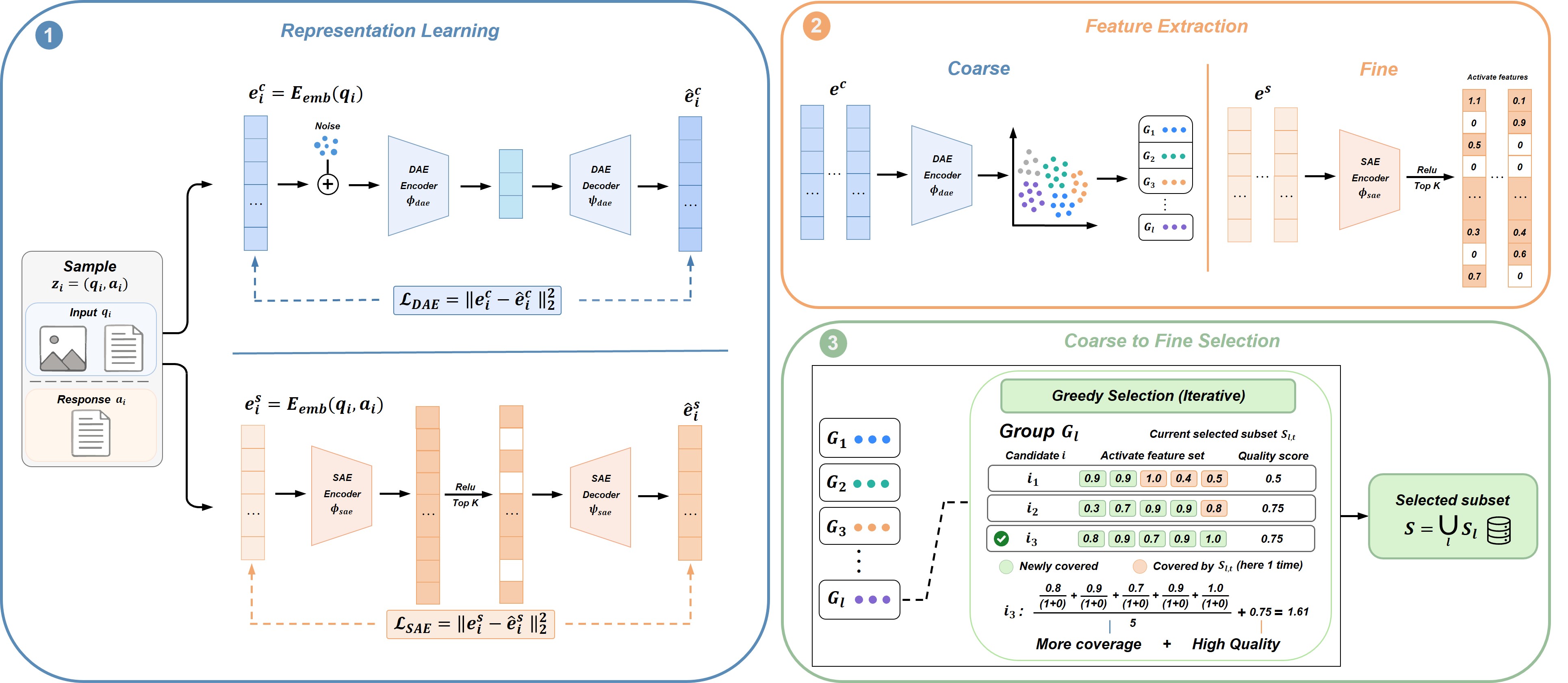}
  \caption{Overview of MASS. MASS first uses a DAE to learn low-dimensional principal manifold coordinates from input side representations and performs coarse grained grouping. It then uses a TopK SAE to extract sparse supervision features from input response representations. Finally, within each group, MASS greedily selects samples by combining feature coverage gain with quality scores, producing the final selected subset.}
  \label{fig:mass_pipeline}
\end{figure*}

\paragraph{DAE for Principal Manifold Encoding.}
We train a DAE in the condition embedding space to map high-dimensional embeddings into stable low-dimensional principal manifold coordinates. Specifically, during training, given $e_i^c$, we construct a perturbation with a fixed magnitude: \begin{equation} u_i = \frac{\epsilon_i}{\|\epsilon_i\|_2},\quad \epsilon_i \sim \mathcal{N}(0,I). \end{equation} We then apply this perturbation to $e_i^c$ and normalize the result as the DAE input: \begin{equation} \widetilde{e}_i^c = \mathrm{Norm}\bigl(e_i^c + \delta u_i\bigr), \end{equation} where $\delta$ is a small fixed constant, and $\mathrm{Norm}(\cdot)$ denotes $L_2$ normalization. The DAE consists of an encoder $\phi_{\mathrm{dae}}$ and a decoder $\psi_{\mathrm{dae}}$. $\phi_{\mathrm{dae}}$ compresses the perturbed high-dimensional representation into a low-dimensional latent representation with dimension 32:
\begin{equation}
h_i = \phi_{\mathrm{dae}}(\widetilde{e}_i^c),
\end{equation}
where $h_i$ is regarded as the low-dimensional principal manifold coordinate of the sample in the condition embedding space. $\psi_{\mathrm{dae}}$ then reconstructs this coordinate back into the original embedding space:
\begin{equation}
\hat{e}_i^c = \mathrm{Norm}\bigl(\psi_{\mathrm{dae}}(h_i)\bigr).
\end{equation}
The training objective is to minimize the squared distance between the original condition embedding and the DAE reconstruction:
\begin{equation}
\mathcal{L}_{\mathrm{DAE}}
=
\frac{1}{N}\sum_{i=1}^{N}
\left\|
e_i^c
-
\hat{e}_i^c
\right\|_2^2 .
\end{equation}

\paragraph{SAE for Sparse Feature Encoding.}
We train a TopK SAE in the supervision embedding space to characterize fine-grained sparse features in complete samples. Unlike the DAE, the SAE takes $e_i^s$ as input, since the effective training signal of a sample comes not only from the input $q_i$, but also from the response $a_i$. Specifically, the SAE consists of an encoder $\phi_{\mathrm{sae}}$ and a decoder $\psi_{\mathrm{sae}}$. Given the supervision representation $e_i^s$, $\phi_{\mathrm{sae}}$ first produces an overcomplete feature preactivation, which is then passed through a ReLU nonlinearity to obtain nonnegative activations:
\begin{equation}
r_i = \mathrm{ReLU}\bigl(\phi_{\mathrm{sae}}(e_i^s)\bigr).
\end{equation}
We then apply the TopK operation for explicit sparsification:
\begin{equation}
k_i = \mathrm{TopK}(r_i, K),
\end{equation}
where $k_i$ is a sparse vector with at most $K$ nonzero activations. In the main experiments, we set the dimension of nonnegative feature activations to 131072 and use $K=64$ for the TopK operation. $\psi_{\mathrm{sae}}$ then reconstructs the supervision embedding from the sparse vector:
\begin{equation}
\hat{e}_i^s = \mathrm{Norm}(\psi_{\mathrm{sae}}(k_i)).
\end{equation}
The training objective is to minimize the squared distance between the original supervision embedding and the SAE reconstruction:
\begin{equation}
\mathcal{L}_{\mathrm{SAE}}
=
\frac{1}{N}\sum_{i=1}^{N}
\left\|
e_i^s - \hat{e}_i^s
\right\|_2^2 .
\end{equation}

\begin{figure*}[t]
  \centering
  \includegraphics[width=0.95\linewidth]{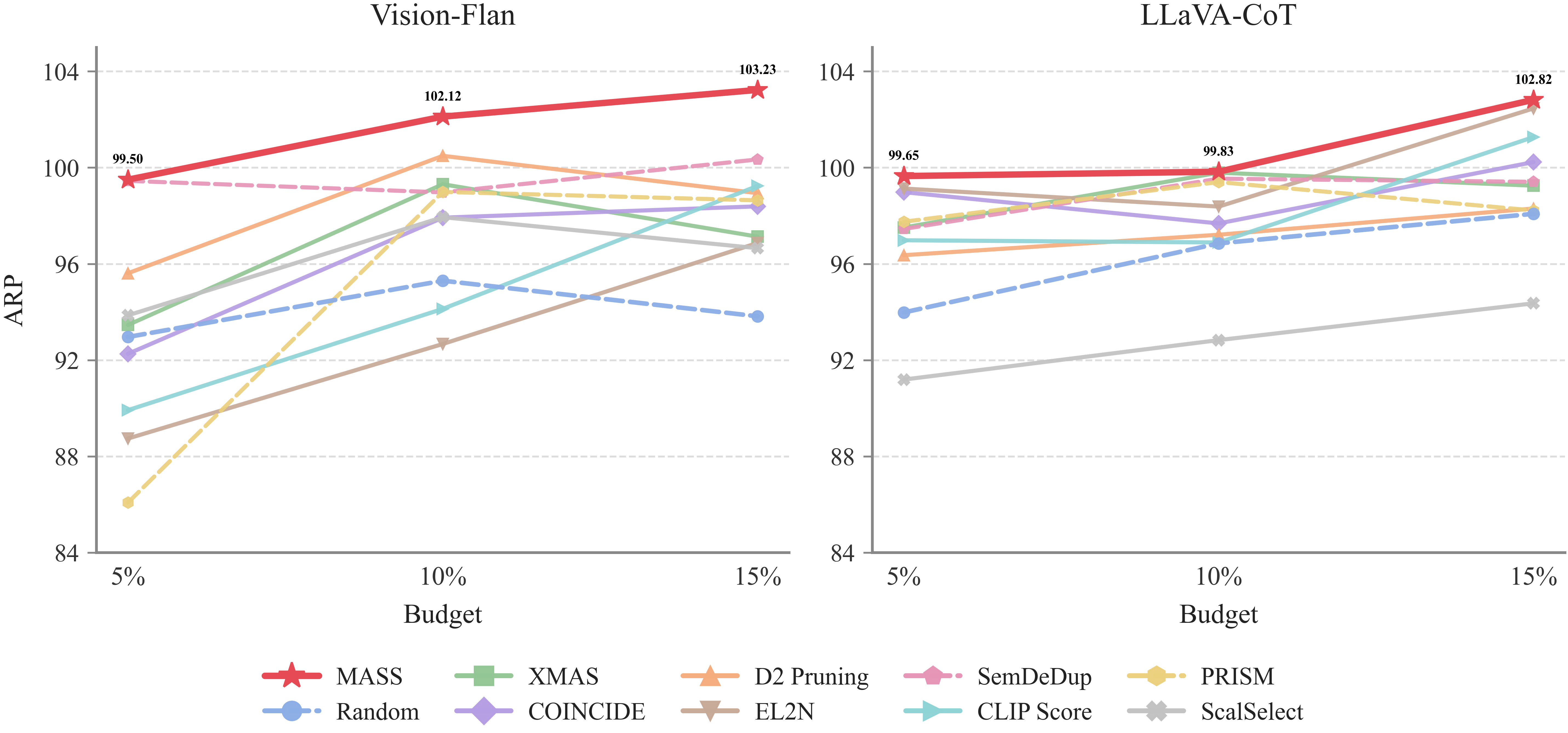}
  \caption{Main results on Vision-Flan and LLaVA-CoT under different data budgets. We report the average relative performance (ARP) of MASS and representative baselines under 5\%, 10\%, and 15\% budgets. MASS achieves consistently strong performance across budgets, outperforming existing baselines on both datasets and surpassing full-data training in several settings.}
  \label{fig:mass_mainresult}
\end{figure*}

\paragraph{Coarse to Fine Selection.}
After training, we first extract DAE latents from clean condition embeddings and perform KMeans clustering, obtaining coarse-grained groups $\{{G}_{l}\}$. For a selection ratio $\rho$, each group ${G}_{l}$ is assigned a sampling budget:
\begin{equation}
B_{l}=\lfloor \rho |{G}_{l}|\rfloor .
\end{equation}
We then sequentially select samples within each ${G}_{l}$. For sample $i$, we define its activated feature set as
\begin{equation}
\Omega_i = \{j: k_{ij}>0\}.
\end{equation}
At selection step $t$, let $S_{l,t}\subseteq {G}_{l}$ denote the current selected subset. The SAE coverage gain of candidate sample $i$ is defined as
\begin{equation}
\Delta_{\mathrm{SAE}}(i\mid S_{l,t})
=
\frac{1}{|\Omega_i|}
\sum_{j\in\Omega_i}
\frac{k_{ij}}{M_j+\epsilon}
\cdot
\frac{1}{n_{l,t}(j)+1},
\end{equation}
where
\begin{equation}
\begin{aligned}
M_j &= \max_i k_{ij},\\
n_{l,t}(j) &= \sum_{m\in S_{l,t}} \mathbf{1}\{j\in\Omega_m\}.
\end{aligned}
\end{equation}
Here, $M_j$ denotes the maximum observed activation of SAE feature $j$ and is used to normalize feature scales, while $n_{l,t}(j)$ counts how many times feature $j$ has been covered in the current subset. We combine the SAE coverage gain with an external quality score $\eta_i\in[0,1]$ and greedily select the highest scoring remaining candidate in group ${G}_{l}$:
\begin{equation}
i_t^{(l)}
=
\arg\max_{i\in{G}_{l}\setminus S_{l,t}}
\left(
\Delta_{\mathrm{SAE}}(i\mid S_{l,t}) + \eta_i
\right).
\end{equation}
The selected sample is then added to the current subset:
\begin{equation}
S_{l,t+1}=S_{l,t}\cup\{i_t^{(l)}\}.
\end{equation}
This greedy procedure is repeated until $|S_{l,t}|=B_{l}$, yielding the selected subset $S_{l}$ for group $l$. The final output set is
\begin{equation}
S = \bigcup_{l} S_{l}.
\end{equation}

\section{Experiments}

\subsection{Experimental Setup}

\paragraph{Datasets.}
We evaluate MASS on Vision-Flan~\cite{xu2024vision} for general instruction tasks and LLaVA-CoT~\cite{xu2025llava} for reasoning tasks, covering different task complexities and data distributions. Dataset details and preprocessing are provided in Appendix~\ref{app:datasets}.

\paragraph{Target Models.}
We use LLaVA-V1.5-7B~\cite{liu2024improved} as the target model on Vision-Flan and Llama-3.2-11B-Vision-Instruct~\cite{grattafiori2024llama} on LLaVA-CoT.
Detailed training hyperparameters are provided in Appendix~\ref{app:training_hyperparameters}.

\paragraph{Data Budgets.}
We evaluate MASS under three data budgets, 5\%, 10\%, and 15\%, for both datasets.

\paragraph{Baselines.}
We compare MASS against 9 data selection
baselines, including Random Selection, XMAS~\cite{naharas2025data}, COINCIDE~\cite{lee2024concept}, SemDeDup~\cite{abbas2023semdedup}, D2 Pruning~\cite{maharana2023d2}, PRISM~\cite{bi2025prism}, ScalSelect~\cite{wu2026scalselect}, CLIP Score~\cite{hessel2021clipscore}, and EL2N~\cite{paul2021deep}. These baselines cover representative data selection paradigms based on importance estimation, diversity distribution, and the combination of both perspectives.

\paragraph{Quality and Embedding Models.}
To reduce computational overhead, we perform scoring with vLLM~\cite{kwon2023efficient} using lightweight models. Specifically, we use Qwen3-VL-4B-Instruct~\cite{bai2025qwen3} for quality scoring on Vision-Flan and Qwen3.5-9B~\cite{team2026qwen3} on LLaVA-CoT. For embedding extraction, we use Qwen3-VL-Embedding-2B~\cite{li2026qwen3} for both datasets.

\paragraph{Evaluation.}
Following the intended use of each dataset, we evaluate the trained target models on different benchmark suites. Vision-Flan and LLaVA-CoT are each evaluated on 12 benchmarks: Vision-Flan focuses on general capability evaluation, while LLaVA-CoT covers both reasoning and general capabilities. The detailed benchmark lists and full results are provided in Appendix~\ref{app:Evaluation}. To ensure stable results, we run all experiments with three random seeds, 0, 42, and 99, and report the average performance. To normalize scores across benchmarks and datasets, we report Average Relative Performance (ARP):
\begin{equation}
\mathrm{ARP} =
\frac{\text{Subset Data Performance}}{\text{Full Data Performance}}
\times 100.
\end{equation}

\subsection{Main Results}

Figure~\ref{fig:mass_mainresult} presents the results of MASS on the Vision-Flan and LLaVA-CoT datasets.

On Vision-Flan, MASS achieves ARP of 99.50, 102.12, and 103.23 under the 5\%, 10\%, and 15\% budgets, respectively, outperforming all baselines. In particular, under the 10\% and 15\% budgets, MASS not only surpasses the strongest baseline but also exceeds Full Data by 2.12 and 3.23 points, respectively. On LLaVA-CoT, MASS also achieves the best performance across all budget settings, with ARP of 99.65, 99.83, and 102.82. It is worth noting that several baselines on this dataset already approach or even surpass Full Data. For example, EL2N reaches 102.46 under the 15\% budget, while MASS further improves upon it and obtains the highest performance.

Overall, MASS consistently achieves the best results on both datasets and across all three data budgets, demonstrating its stable and effective data selection capability under different data distributions and task complexities. These results validate our central hypothesis: effective data selection should not be restricted to global sampling in the original embedding space. Instead, it should be formulated as a hierarchical coverage problem that preserves the principal  manifold distribution of the data while also covering locally fine grained features.

\section{Analysis and Ablation Studies}
\subsection{Effect of DAE Manifold Coordinates}
To verify whether coarse grouping requires DAE manifold coordinates, we compare MASS with a variant on Vision-Flan under the 10\% budget, where DAE clustering is replaced by clustering in the original embedding space. All other settings are kept unchanged.

As shown in Table~\ref{tab:5_1}, clustering based on raw embeddings achieves an ARP of 101.59, whereas clustering based on the DAE principal  manifold coordinates achieves a higher ARP of 102.12. This result suggests that although raw embeddings contain rich semantic information, their geometric structure may still mix redundant dimensions, local perturbations, and non dominant variations, leading to less stable coarse grained grouping. In contrast, the low-dimensional principal  manifold coordinates learned by DAE can more effectively capture the principal semantic directions of the data, providing more reliable coarse grained region partitions for subsequent SAE based coverage selection within each cluster and thereby improving the final performance.

\subsection{Complementarity of DAE and SAE}

To analyze the complementarity of DAE and SAE, we construct two variants on Vision-Flan under the 10\% budget. The first removes SAE and uses only DAE clustering followed by random sampling within each cluster. The second removes DAE and directly performs fine granularity coverage selection based on SAE features.

\begin{figure}[t]
  \centering
  \includegraphics[width=0.8\linewidth]{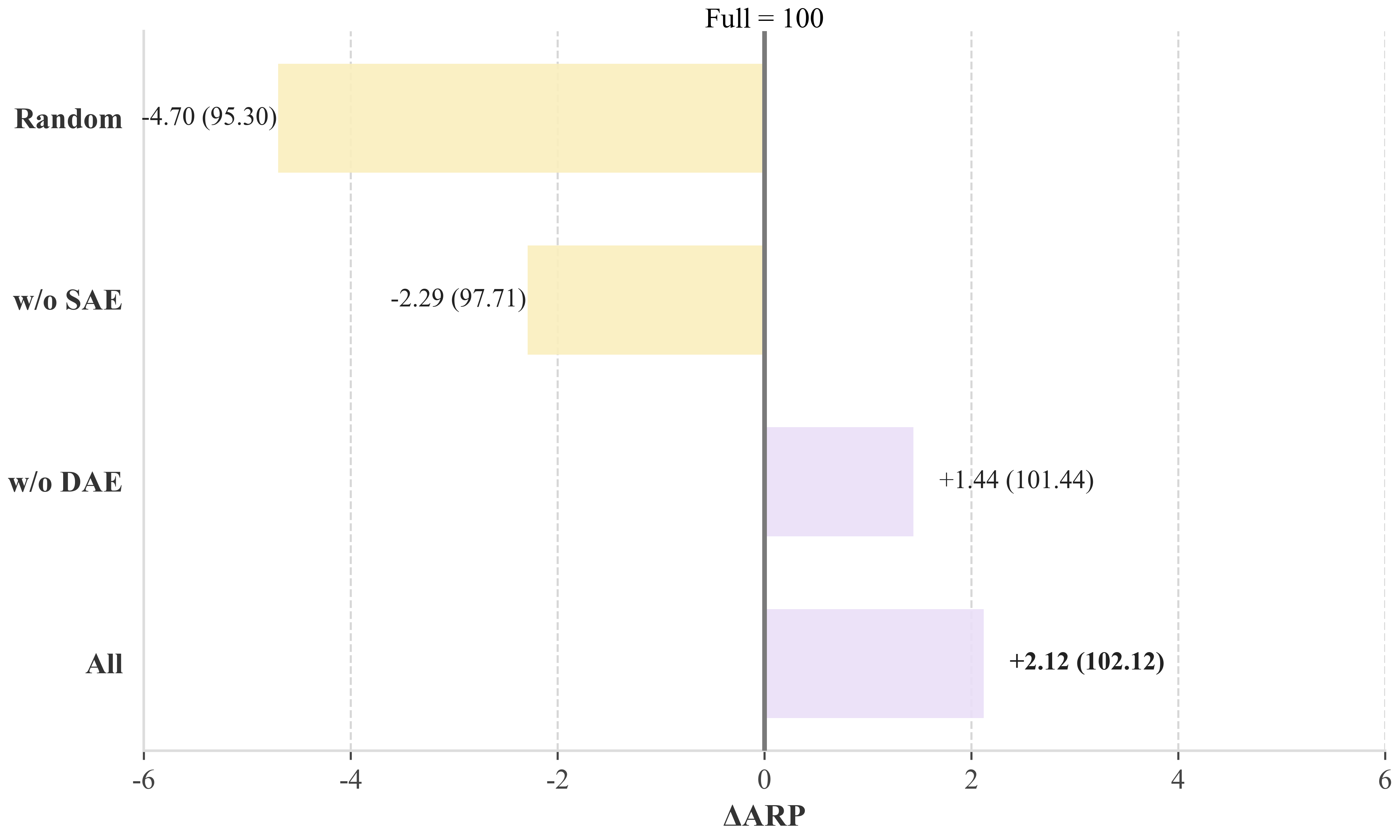}
  \caption{Complementarity of DAE and SAE. DAE alone improves over random selection by preserving coarse manifold coverage, while SAE alone brings a larger gain through fine grained feature coverage. Combining both components yields the best performance, demonstrating the complementarity between DAE and SAE}
  \label{fig:5_2}
\end{figure}

As shown in Figure~\ref{fig:5_2}, random selection obtains an ARP of 95.30, while the variant using only DAE reaches 97.71, indicating that principal manifold grouping provides useful coverage at a coarse granularity. The variant using only SAE achieves an ARP of 101.44, showing that fine granularity feature coverage can effectively improve the quality of sample coverage. The full MASS achieves the best ARP of 102.12, outperforming both single module variants. These results demonstrate that DAE and SAE are complementary at two levels of coverage: DAE maintains coarse granularity coverage over the principal manifold, while SAE enhances fine granularity feature coverage within each manifold region.

\begin{table}[t]
\centering

\setlength{\tabcolsep}{18pt}
\begin{tabular}{lc}
\toprule
Method & ARP \\
\midrule
Raw & 101.59 \\
DAE & \textbf{102.12} \\
\bottomrule
\end{tabular}
\caption{Effect of DAE manifold coordinates. DAE clustering improves ARP over raw embedding clustering, showing the benefit of principal manifold coordinates for coarse grouping.}
\label{tab:5_1}
\end{table}

\subsection{Sensitivity to DAE Configuration}

To analyze the effect of DAE configuration on MASS, we vary the DAE latent dimension on Vision-Flan under the 10\% budget, while keeping all other settings unchanged.

\begin{figure}[t]
  \centering
  \includegraphics[width=0.85\linewidth]{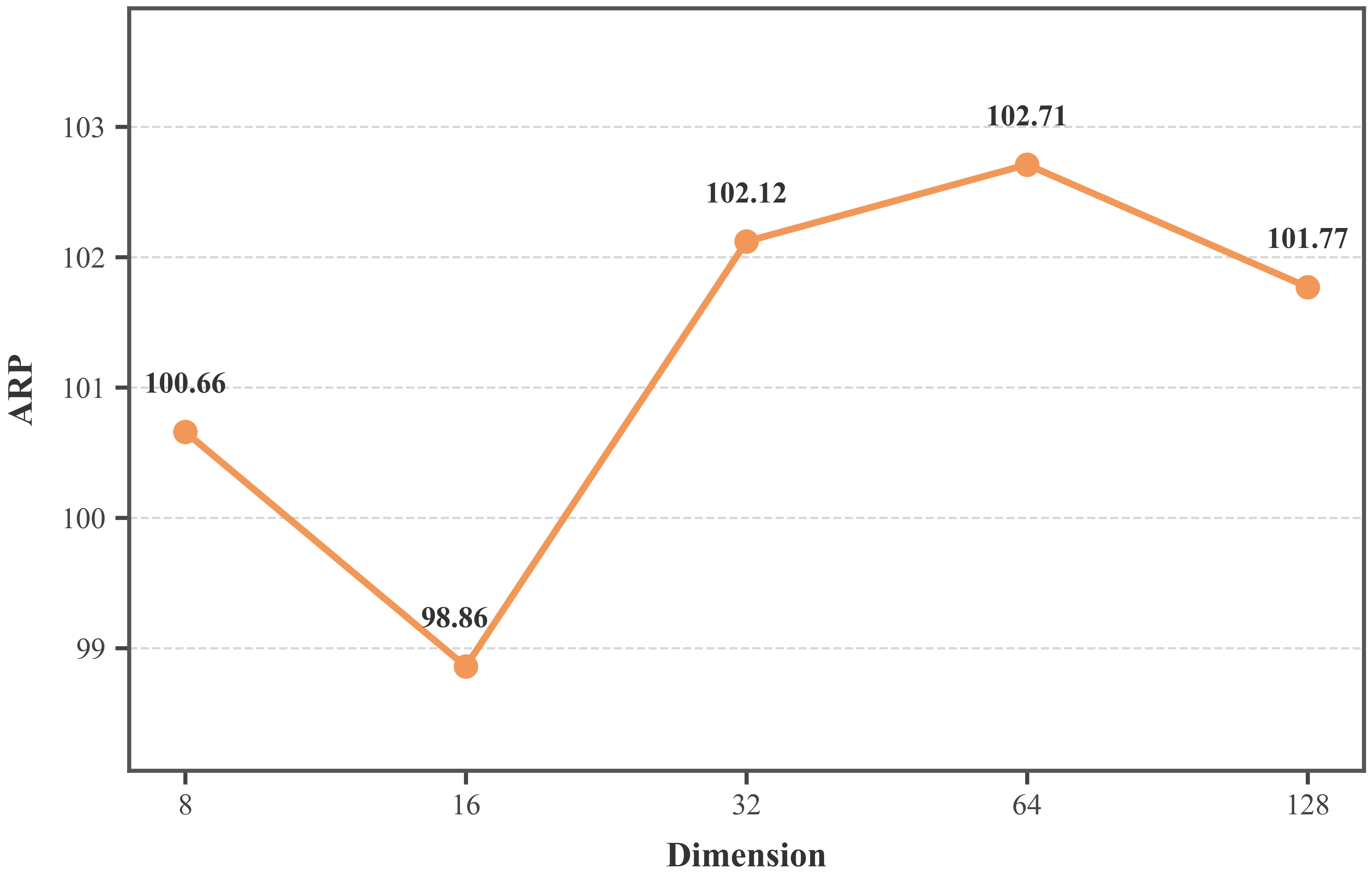}
  \caption{Sensitivity to DAE configuration. Performance drops when the latent dimension is either too small or too large, while a moderate dimension of 64 achieves the best ARP, suggesting the importance of proper latent capacity.}
  \label{fig:5_3}
\end{figure}

As shown in Figure~\ref{fig:5_3}, when the latent dimension is set to 8, 16, 32, 64, and 128, MASS achieves ARP of 100.66, 98.86, 102.12, 102.71, and 101.77, respectively. The results show that either too small or too large a DAE latent dimension can hurt the final performance. When the dimension is too small, the principal manifold coordinates may fail to preserve sufficient semantic structure, leading to inadequate coarse granularity grouping. When the dimension is too large, the embedding space may reintroduce redundant dimensions and local perturbations, weakening the ability of DAE to extract the principal semantic directions. Overall, the 64 dimensional setting achieves the best performance, suggesting that a moderate compression dimension better balances principal manifold structure preservation and suppression of non dominant variations.

\subsection{Sensitivity to SAE Configuration}

To analyze the effect of SAE configuration on MASS, we study both the SAE feature dimension and the TopK sparsity on Vision-Flan. For the feature dimension analysis, we compare SAE dimensions of 65536 and 131072 under 5\%, 10\%, and 15\% budgets. For the TopK analysis, we fix the SAE feature dimension to 131072 and vary the TopK value under the 10\% data budget. All other settings are kept unchanged.

\begin{figure}[t]
  \centering
  \includegraphics[width=0.85\linewidth]{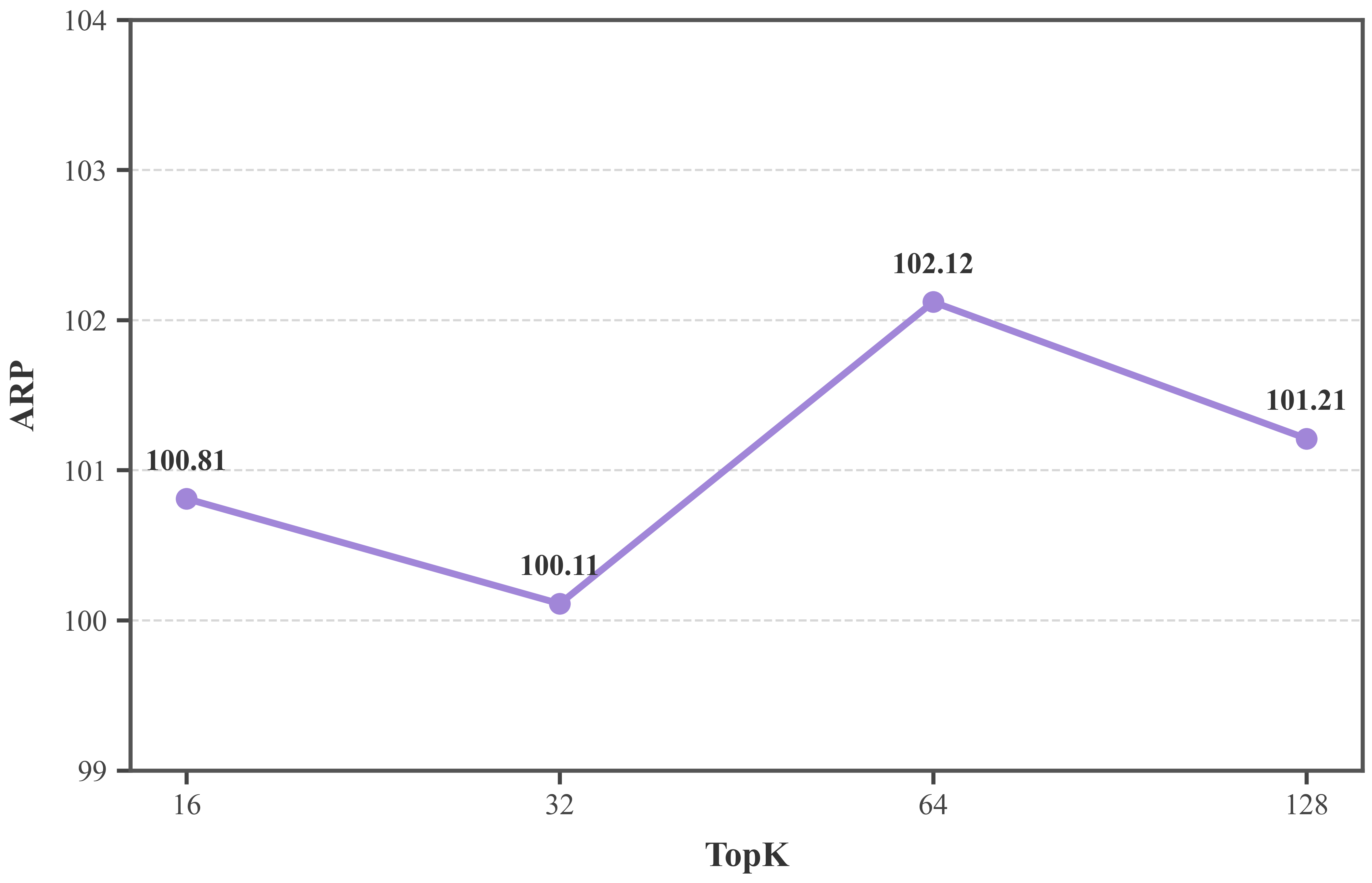}
  \caption{Effect of TopK sparsity. MASS achieves the best ARP with TopK of 64, while both smaller and larger TopK values lead to lower performance, indicating the importance of a proper sparsity level.}
  \label{fig:5_4_1}
\end{figure}

\paragraph{Effect of TopK Sparsity.}
As shown in Figure~\ref{fig:5_4_1}, when TopK is set to 16, 32, 64, and 128, MASS achieves ARP of 100.81, 100.11, 102.12, and 101.21, respectively. These results suggest that a too small TopK limits the number of activated features for each sample, resulting in insufficient fine granularity coverage. In contrast, a too large TopK weakens the sparsity of SAE representations and may introduce more non essential features. TopK of 64 achieves the best result, indicating that a moderate sparsity level better balances feature coverage capacity and representation selectivity.

\begin{figure}[t]
  \centering
  \includegraphics[width=0.85\linewidth]{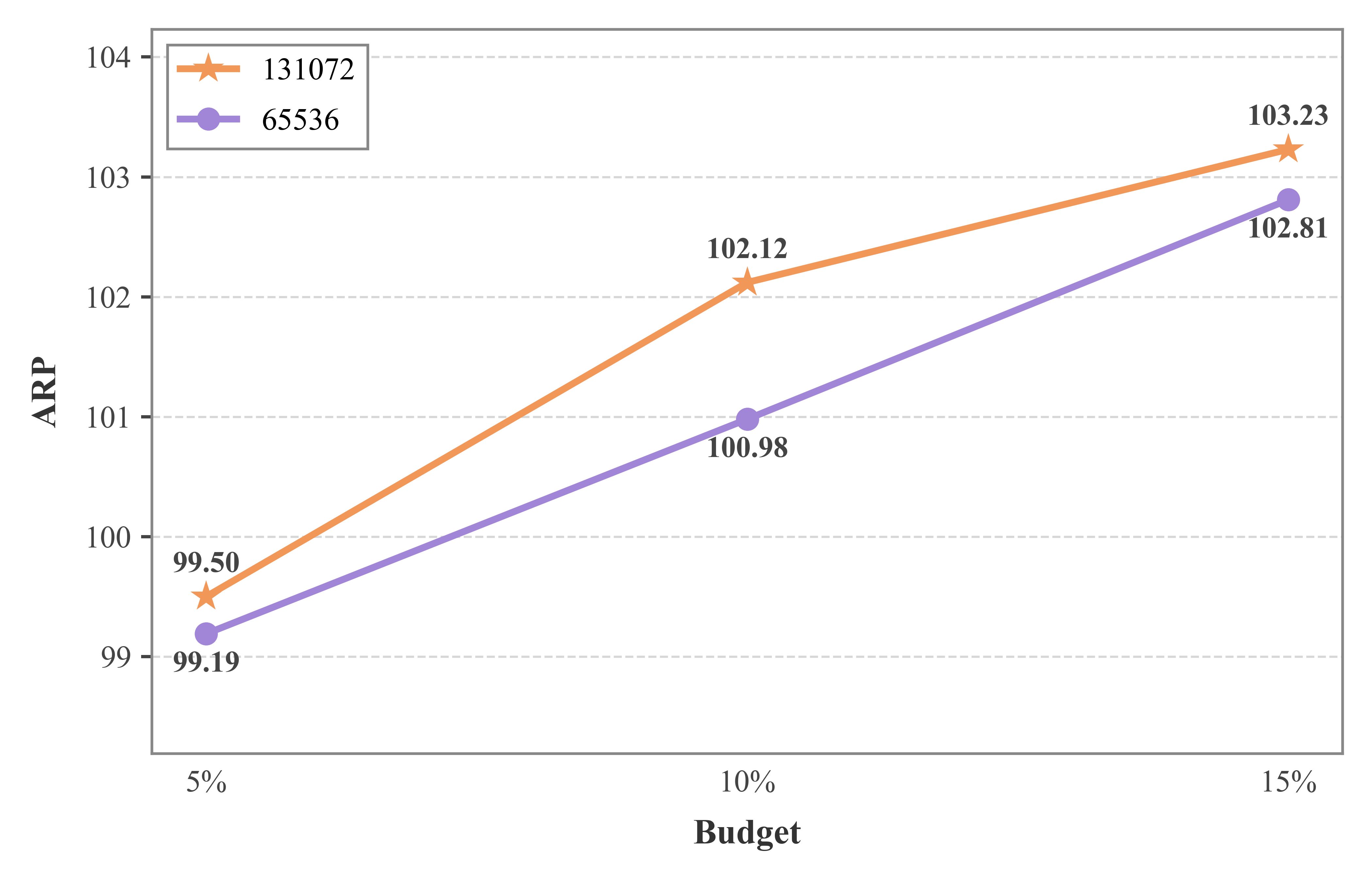}
  \caption{Effect of SAE feature dimension. Increasing the SAE feature dimension from 65536 to 131072 consistently improves ARP across data budgets, suggesting the importance of sufficient sparse feature capacity.}
  \label{fig:5_4_2}
\end{figure}

\paragraph{Effect of SAE Feature Dimension.}
As shown in Figure~\ref{fig:5_4_2}, when the SAE feature dimension is 65536, MASS achieves ARP of 99.19, 100.98, and 102.81 under the 5\%, 10\%, and 15\% budgets, respectively. Increasing the feature dimension to 131072 improves the corresponding ARP to 99.50, 102.12, and 103.23. These results indicate that a larger SAE feature space provides stronger capacity for fine granularity feature representation, thereby improving the effectiveness of within cluster sparse feature coverage.

\subsection{Effect of Quality Signals}

To analyze the effect of external quality scores on MASS, we further test two settings on Vision-Flan under the 10\% data budget: removing the quality score and replacing the scoring model with LLaVA-OneVision-1.5-4B-Instruct~\cite{an2025llava}. All other settings are kept unchanged.

As shown in Table~\ref{tab:5_5}, random selection obtains an ARP of only 95.30, while MASS without quality scores reaches 99.96. This indicates that the hierarchical coverage mechanism based on DAE and SAE already provides a strong data selection effect. After adding external quality scores, MASS achieves 102.12 ARP with the original scoring model and 102.44 ARP with LLaVA-OneVision-1.5-4B-Instruct. These results show that external quality scores can effectively filter low quality samples and complement both coarse granularity principal manifold coverage and fine granularity feature coverage. Moreover, different scoring models bring consistent improvements, suggesting that MASS is not overly sensitive to the specific choice of quality model.

\begin{table}[t]
\centering

\setlength{\tabcolsep}{18pt}
\begin{tabular}{lc}
\toprule
Method & ARP \\
\midrule
Random & 95.30 \\
w/o Quality & 99.96 \\
Original Scorer & 102.12 \\
LLaVA-OneVision & \textbf{102.44} \\
\bottomrule
\end{tabular}
\caption{Effect of quality signals. MASS without quality signals already improves over random selection, while different scoring models further increase ARP, showing that quality signals are useful and robust across scorers.}
\label{tab:5_5}
\end{table}

\begin{table}[t]
\centering

\setlength{\tabcolsep}{18pt}
\begin{tabular}{lc}
\toprule
Method & ARP \\
\midrule
Random & 95.30 \\
GME & \textbf{101.68} \\
\bottomrule
\end{tabular}
\caption{Effect of embedding source. MASS remains effective with GME embeddings, substantially outperforming random selection.}
\label{tab:5_6}
\end{table}

\subsection{Effect of Embedding Source}

To analyze the effect of the embedding source on MASS, we replace the original embedding source with GME~\cite{zhang2024gme} on Vision-Flan under the 10\% data budget, while keeping all other settings unchanged.

As shown in Table~\ref{tab:5_6}, random selection obtains an ARP of 95.30, while MASS with GME embeddings achieves 101.68, substantially outperforming random selection. This result shows that the hierarchical selection mechanism of MASS does not rely on a single embedding source and can still maintain strong performance after changing the embedding representation. Meanwhile, the result with GME is slightly lower than the 102.12 ARP obtained with the original embedding source, which is consistent with the relatively weaker representation ability of GME in our setting. This suggests that MASS has certain adaptability to different embedding sources, while a stronger embedding representation can still provide a better basis for DAE principal manifold grouping and SAE fine granularity feature coverage.

\begin{figure}[t]
  \centering
  \includegraphics[width=0.98\linewidth]{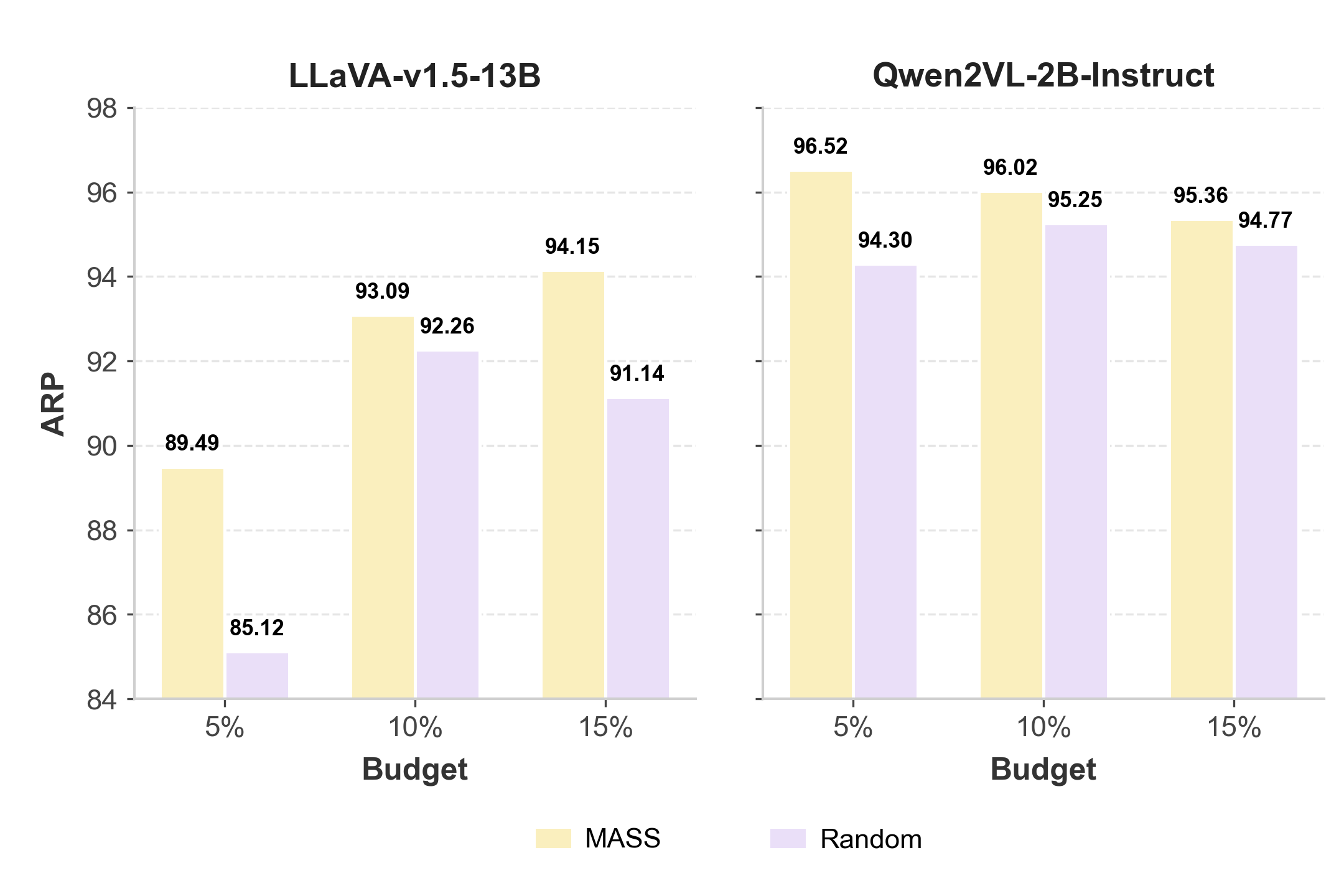}
  \caption{Robustness across target models. We evaluate MASS by replacing the target models with LLaVA-v1.5-13B on Vision-Flan and Qwen2VL-2B-Instruct on LLaVA-CoT, the results show that MASS maintains stable performance across different model scales and model families}
  \label{fig:5_7}
\end{figure}
\subsection{Robustness Across Target Models}

To evaluate the robustness of MASS across target models, we test LLaVA-v1.5-13B~\cite{liu2024improved} on Vision-Flan and Qwen2VL-2B-Instruct~\cite{wang2024qwen2} on LLaVA-CoT, covering both variation in parameter scale within the same model family and variation across different model families. All
other settings are kept unchanged.

As shown in Figure~\ref{fig:5_7}, on Vision-Flan, MASS achieves ARP of 89.49, 93.09, and 94.15 under the 5\%, 10\%, and 15\% budgets, respectively, outperforming random selection with 85.12, 92.26, and 91.14. On LLaVA-CoT, MASS achieves 96.52, 96.02, and 95.36 ARP under the three budgets, also consistently outperforming random selection with 94.30, 95.25, and 94.77. These results show that MASS remains effective under different target model scales and model families, demonstrating its robustness across target models.

\section{Conclusion and Limitations}
\paragraph{Conclusion.}
This paper formulates supervised fine tuning data selection as a coarse to fine hierarchical coverage problem. To address the entanglement of principal semantic directions, fine grained supervision differences, and local noise in the original embedding space, we propose MASS. MASS uses a dense autoencoder to learn low-dimensional principal manifold coordinates for coarse grained grouping, and then performs fine grained selection within each group by combining TopK sparse autoencoder features with quality signals. Experimental results show that MASS consistently outperforms existing baselines across multiple datasets, data budgets, embedding sources, and target models, validating the effectiveness of principal manifold coverage and within group sparse feature coverage. MASS provides a hierarchical coverage paradigm for embedding based data selection.

\paragraph{Limitations.}
Despite its stable and effective performance, MASS has several limitations. First, MASS relies on external embedding source and quality scoring models, so the selected subset may be affected when these external signals are severely biased. Second, MASS focuses on hierarchical coverage selection within a given candidate pool; when the pool itself lacks certain task types or supervision patterns, the method cannot compensate for such fundamental distributional gaps. Finally, MASS does not study how to select subsets from general data collections for specific downstream tasks, leaving its applicability to task oriented data selection for future exploration.

\clearpage
\appendix

\section{Embedding Geometry Analysis}
\label{app:embedding_geometry}

To better understand whether the original embedding space is suitable for direct data selection, we analyze the embedding geometry of Vision-Flan and LLaVA-CoT. Both datasets use the same embedding preprocessing procedure as MASS, where embeddings are first centered and then $L_2$ normalized. Table~\ref{tab:embedding_geometry} summarizes several geometry statistics. The number of PCA components required to explain 90\% of the variance and the PCA effective rank are used to characterize the global linear complexity of the embedding distribution. The TwoNN intrinsic dimension is used to estimate the intrinsic dimensionality of the data. The local PCA dimension estimates, for each sample neighborhood, the minimum number of local principal components required to explain 90\% of the local variance. Finally, we compare the neighborhood preservation of PCA and Isomap to measure how well low-dimensional representations preserve the local neighborhood structure of the original embedding space.

The results show that the embedding distributions of both datasets have high global linear complexity. Although the embedding dimension of both Vision-Flan and LLaVA-CoT is 2048, Vision-Flan still requires 379 PCA components to explain 90\% of the variance, while LLaVA-CoT requires 459 components. Meanwhile, their PCA effective ranks reach 255.06 and 275.17, respectively. These results indicate that the data are not concentrated in a simple low-dimensional linear subspace, and that global distances and aggregation patterns in the original embedding space may be affected by many variation directions.

At the same time, the intrinsic dimension and local PCA results show that the data exhibit much lower effective dimensionality in local neighborhoods. The TwoNN intrinsic dimension of Vision-Flan is 8.25, and its mean and median local PCA dimensions are 17.38 and 19, respectively. The TwoNN intrinsic dimension of LLaVA-CoT is 11.88, and its mean and median local PCA dimensions are 20.88 and 21, respectively. The large gap between global PCA complexity and local dimensionality suggests that instruction data embeddings are not unstructured high-dimensional point clouds. Instead, they are better characterized as globally complex distributions with locally low-dimensional geometry.

Furthermore, the comparison between PCA and Isomap neighborhood preservation indicates that this local structure is nonlinear. On Vision-Flan, Isomap achieves a neighborhood preservation score of 0.499, which is higher than the PCA score of 0.383. On LLaVA-CoT, Isomap also improves the score from 0.281 under PCA to 0.380. Since PCA captures a global linear projection while Isomap better preserves nonlinear neighborhood geometry, this result suggests that local neighborhood relations in the embedding space are more consistent with a nonlinear manifold structure than with a single global linear structure.

These observations support the central motivation of this work. The original embedding space of complex instruction data contains global semantic variation, local fine grained differences, and task irrelevant perturbations at the same time. If data selection is performed directly in this space, a single distance or diversity criterion may fail to distinguish these different levels of variation. Based on this observation, MASS formulates data selection as a coarse to fine hierarchical coverage process. It first uses a DAE to learn stable low-dimensional principal manifold coordinates for constructing coarse grained semantic regions, and then uses an SAE to extract sparse supervision features for fine grained coverage selection within each region.

\section{Implementation Details of MASS}
\label{app:data_dpo_implementation}

This section supplements the implementation details and hyperparameters omitted from Section~\ref{subsec:detail_mass} to ensure experimental reproducibility.

\paragraph{Dual View Encoding.}
For each sample $z_i=(q_i,a_i)$, the condition embedding $e_i^c$ is obtained by feeding only $q_i$ into the frozen embedding model, while the supervision embedding $e_i^s$ is obtained by feeding the concatenated pair $(q_i,a_i)$ into the embedding model.

\paragraph{Dense Autoencoder for Principal Manifold Encoding.}
The dense autoencoder is used to learn low-dimensional principal manifold coordinates from input side condition embeddings. During training, input embeddings are first $L_2$ normalized. We then add spherical noise with a fixed magnitude of 0.03 and apply $L_2$ normalization again before feeding them into the model. The model is a simple MLP autoencoder. Its encoder has hidden dimensions of 1024, 512, and 128 with GELU activations, and maps each input to a low-dimensional latent space. The decoder uses a symmetric architecture to reconstruct the latent representation back to the original embedding space. In main experiments, latent dimension is set to 32. We train the dense autoencoder with AdamW using a learning rate of $1\times10^{-4}$, a weight decay of $1\times10^{-4}$, a batch size of 128, and 200 epochs. We use 2\% of data for validation and select the best checkpoint based on validation reconstruction loss. After training, clean condition embeddings are encoded into latent coordinates, and KMeans clustering with 256 clusters is applied for coarse grained grouping.

\paragraph{Sparse Autoencoder for Sparse Feature Encoding.}
The sparse autoencoder is used to model fine grained sparse features in the supervision embeddings formed jointly by the input and the response. Before training, we compute the training set mean, center supervision embeddings, and apply $L_2$ normalization. We use a TopK sparse autoencoder. Its encoder maps each input into an overcomplete feature space, applies ReLU, and keeps the TopK largest activations. The decoder reconstructs the original supervision embedding from these sparse activations and normalizes its output. In main experiments, the sparse autoencoder uses 131072 features with TopK set to 64. We train it with AdamW using a learning rate of $1\times10^{-4}$, zero weight decay, a batch size of 128, and 100 epochs. We use 5\% of data for validation and select the best checkpoint based on validation reconstruction loss.

\begin{figure*}[t]
    \centering
    \includegraphics[width=0.5\linewidth]{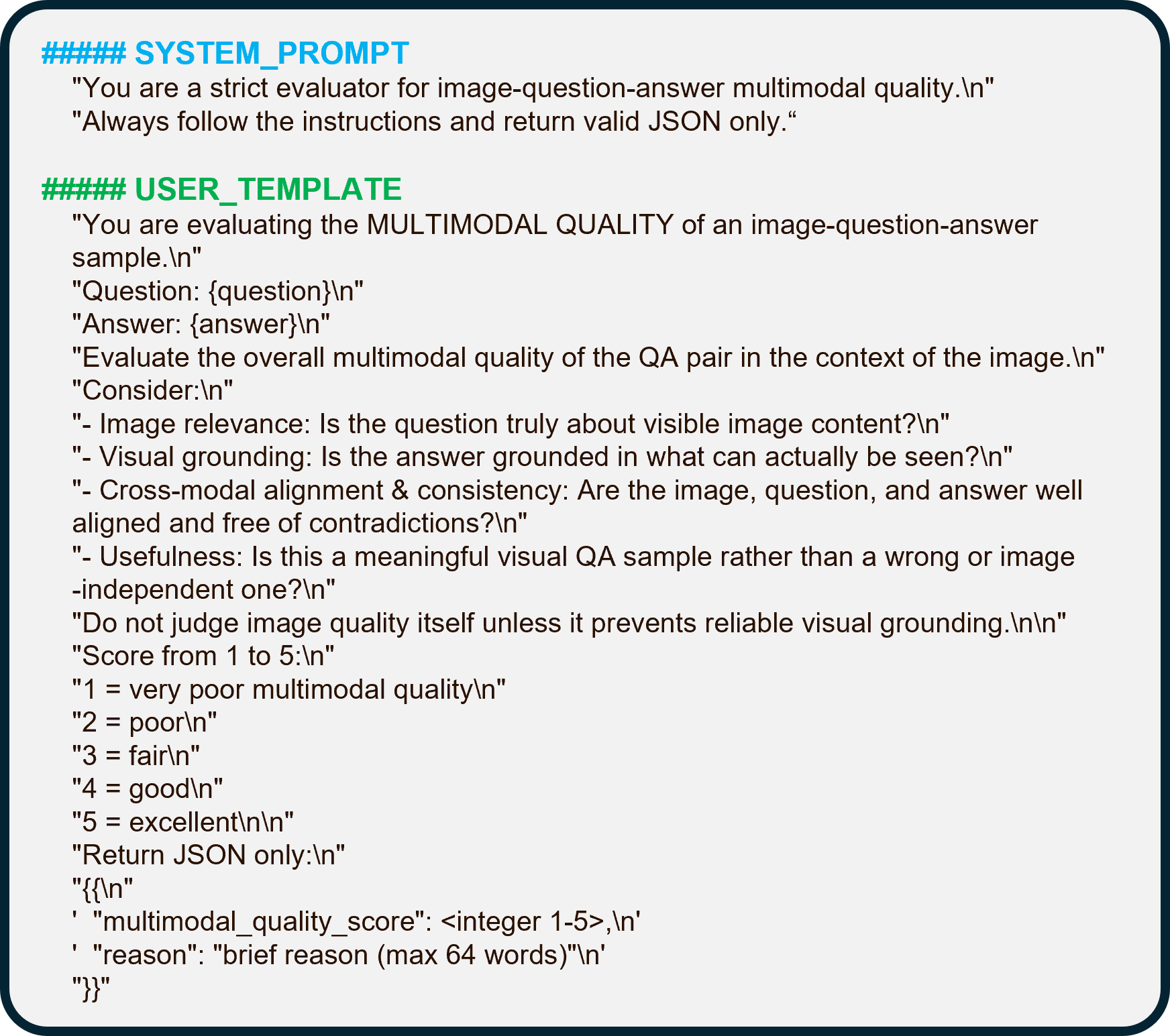}
    \caption{Scoring prompt for estimating the quality score on Vision-Flan.}
    \label{fig:prompt_vf_mm}
\end{figure*}

\begin{figure*}[t]
    \centering
    \includegraphics[width=0.5\linewidth]{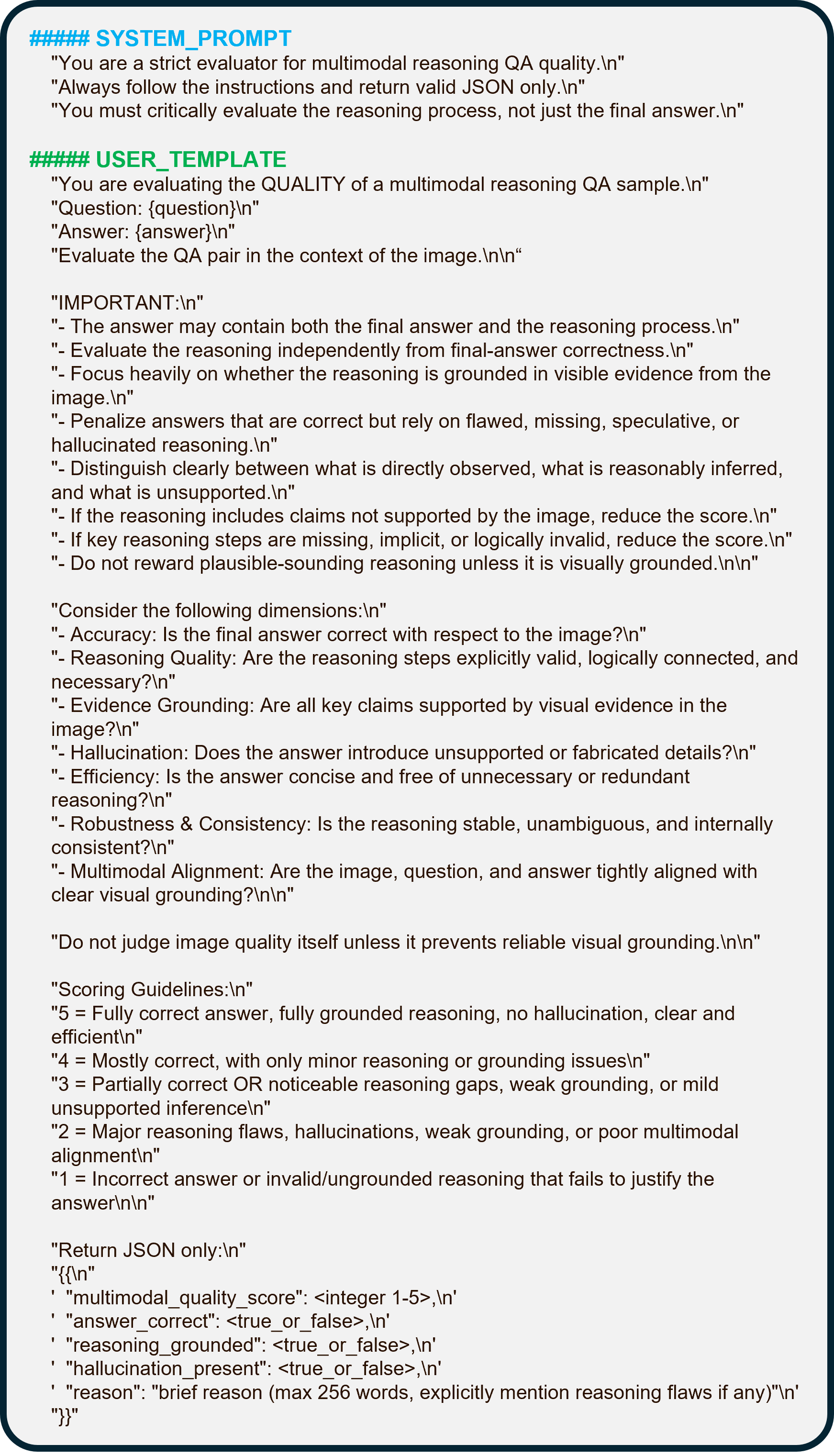}
    \caption{Scoring prompt for estimating the quality score on LLaVA-CoT.}
    \label{fig:prompt_llavacot}
\end{figure*}

\paragraph{Coarse to Fine Selection.}
The external quality score is obtained by feeding each sample into a multimodal model. For each sample, we use a discrete rating scale from 1 to 5, and normalize the rating to $[0,1]$. Specifically, the scores $1,2,3,4,5$ are mapped to $0,0.25,0.5,0.75,1$, respectively. The prompts used for quality scoring are shown in Figure~\ref{fig:prompt_vf_mm} and Figure~\ref{fig:prompt_llavacot}.

\section{Datasets and Preprocessing Details}
\label{app:datasets}
\subsection{Datasets}
\paragraph{Vision-Flan.}
Vision-Flan~\cite{xu2024vision} is a high-quality human-annotated visual instruction tuning dataset designed to improve the generalization ability of vision-language models across diverse visual tasks. It is constructed from publicly available academic vision datasets and covers 187 fine-grained visual tasks, each paired with expert-written and carefully validated task instructions. The dataset reduces synthetic bias and hallucination risk while providing broad task coverage, containing approximately 1000 samples per task and about 186K samples in total.
\paragraph{LLaVA-CoT.}
LLaVA-CoT-100k~\cite{xu2025llava} is an image-text reasoning instruction dataset designed to improve the structured reasoning ability of vision-language models. It integrates approximately 99K image-question-answer pairs from a mixture of general-purpose and science-targeted VQA datasets. Each sample is annotated with a structured reasoning response consisting of summary, caption, reasoning, and conclusion stages.

\subsection{Preprocessing Procedures}
For Vision-Flan, we use the original dataset without additional filtering or format conversion. For LLaVA-CoT, we remove samples without associated images, resulting in 98572 image-text reasoning samples. The reasoning process and final answer in each response are wrapped with \verb|<think></think>| and \verb|<answer></answer>|, respectively.

\section{Training Hyperparameters}
\label{app:training_hyperparameters}

Table~\ref{tab:training_hyperparameters} summarizes the training hyperparameters used for all target models. 
For LLaVA-V1.5-7B and LLaVA-V1.5-13B, we follow the hyperparameter settings from the official LLaVA training code. For Qwen2-VL-2B-Instruct and Llama-3.2-11B-Vision-Instruct, since both models are already instruction-tuned checkpoints, we use a conservative learning rate of $1\times10^{-5}$ for full-parameter fine-tuning. 
We set the batch size to 64 for these two models because using a batch size of 128 leads to too few optimization steps within one epoch, both for the full dataset and for subset training, making the training insufficient.

\section{Evaluation}
\label{app:Evaluation}

\subsection{Evaluation Benchmarks}
\label{sec:benchmarks}

We choose benchmarks according to the modality and target capability of each dataset. For Vision-Flan, we use benchmarks covering general multimodal understanding, visual question answering, OCR, document understanding, chart understanding, and science-diagram reasoning. For LLaVA-CoT, we combine multimodal mathematical and logical reasoning benchmarks with general vision-language benchmarks to evaluate both reasoning ability and overall multimodal robustness. For Qwen2-VL-2B-Instruct trained on LLaVA-CoT, we observe that full-data training still yields extremely poor performance on We-Math~\cite{qiao2025we}, LogicVista~\cite{xiao2024logicvista}, DynaMath~\cite{zou2024dynamath}, and MMStar~\cite{chen2024we}, making these benchmarks less informative for comparing data selection methods under this target model. Therefore, for this model we evaluate on the remaining eight benchmarks. The full benchmark list is shown in Table~\ref{tab:benchmark_list}.

\subsection{Detailed Evaluation Results}
\label{sec:detailed_results}

For the main experiments, Tables~\ref{tab:full_comparison_vf_mass} and~\ref{tab:full_comparison_llavacot_mass} present the full benchmark-level comparison results on Vision-Flan and LLaVA-CoT, respectively, covering MASS and all data selection baselines. We further provide benchmark-level results for the analysis and ablation studies discussed in the main paper and appendix. Specifically, Table~\ref{tab:dae_manifold_ablation_vf} reports the ablation results of DAE manifold coordinates. Table~\ref{tab:dae_sae_complementarity_vf} analyzes the complementarity between DAE and SAE. Table~\ref{tab:dae_latent_dim_sensitivity_vf} reports the sensitivity analysis of the DAE latent dimension. Table~\ref{tab:sae_configuration_sensitivity_vf} reports the sensitivity analysis of the SAE feature dimension and TopK sparsity. Table~\ref{tab:quality_signal_ablation_vf} reports the results under different quality signal sources. Table~\ref{tab:embedding_source_ablation_vf} reports the results under different embedding sources. Finally, Table~\ref{tab:robustness_vf_llava15_13b} shows the Vision-Flan results with LLaVA-V1.5-13B as the target model, while Table~\ref{tab:robustness_llavacot_qwen2vl_2b} shows the LLaVA-CoT results with Qwen2-VL-2B-Instruct.

\section{Time Cost Analysis}

We compare the time cost of different data selection methods
on LLaVA-CoT. Table~\ref{tab:time_cost} reports the GPU hours
required by each method on a single NVIDIA A6000. MASS takes
approximately 15.39 GPU hours in total. Although its computational
cost is slightly higher than that of some baselines, most of the
time is spent on quality scoring and embedding extraction, and
these results can be reused in other data usage scenarios, such as
subsequent curriculum learning. In addition, compared with the
cost of full SFT training, this preprocessing overhead remains
acceptable. As shown by the main experimental results, MASS
achieves the best performance under multiple data budgets,
demonstrating a favorable balance between selection effectiveness
and computational cost.

\clearpage

\begin{table*}[t]
\centering
\begin{tabular}{lcc}
\toprule
Metric & Vision-Flan & LLaVA-CoT \\
\midrule
Embedding dimension & 2048 & 2048 \\
PCA components for 90\% variance & 379 & 459 \\
PCA effective rank & 255.06 & 275.17 \\
TwoNN intrinsic dimension & 8.25 & 11.88 \\
Local PCA dimension (mean) & 17.38 & 20.88 \\
Local PCA dimension (median) & 19 & 21 \\
PCA neighborhood preservation & 0.383 & 0.281 \\
Isomap neighborhood preservation & 0.499 & 0.380 \\
\bottomrule
\end{tabular}
\caption{Embedding geometry statistics of Vision-Flan and LLaVA-CoT. Both datasets exhibit globally complex but locally low-dimensional nonlinear structures.}
\label{tab:embedding_geometry}
\end{table*}

\begin{table*}[t]
\centering
\setlength{\tabcolsep}{3.5pt}
\resizebox{0.95\textwidth}{!}{
\begin{tabular}{lccccccccc}
\toprule
\textbf{Model} & \textbf{Strategy} & \textbf{Epoch} & \textbf{Batch size} & \textbf{LR} & \textbf{Optimizer} & \textbf{LR scheduler} & \textbf{Warmup ratio} & \textbf{LoRA rank} & \textbf{LoRA alpha} \\
\midrule
LLaVA-V1.5-7B & LoRA & 1 & 128 & $2\times10^{-4}$ & AdamW & Cosine & 0.03 & 128 & 256 \\
LLaVA-V1.5-13B & LoRA & 1 & 128 & $2\times10^{-4}$ & AdamW & Cosine & 0.03 & 128 & 256 \\
Qwen2-VL-2B-Instruct & Full & 1 & 64 & $1\times10^{-5}$ & AdamW & Cosine & 0.03 & -- & -- \\
Llama-3.2-11B-Vision-Instruct & Full & 1 & 64 & $1\times10^{-5}$ & AdamW & Cosine & 0.03 & -- & -- \\
\bottomrule
\end{tabular}
}

\caption{Training hyperparameters for target models.}
\label{tab:training_hyperparameters}
\end{table*}

\begin{table*}[t]
\centering
\begin{tabular}{cl}
\toprule
\textbf{Dataset} & \textbf{Benchmark} \\
\midrule
\multirow[c]{12}{*}{Vision-Flan}
& GQA~\cite{hudson2019gqa} \\
& VizWiz~\cite{gurari2018vizwiz} \\
& TextVQA~\cite{singh2019towards} \\
& ScienceQA-IMG~\cite{saikh2022scienceqa} \\
& MME~\cite{liang2024survey} \\
& MMBench~\cite{liu2024mmbench} \\
& AI2D~\cite{kembhavi2016diagram} \\
& ChartQA~\cite{masry2022chartqa} \\
& DocVQA~\cite{mathew2021docvqa} \\
& InfoVQA~\cite{mathew2022infographicvqa} \\
& MMStar~\cite{chen2024we} \\
& OCRBench~\cite{liu2024ocrbench} \\
\midrule

\multirow[c]{12}{*}{LLaVA-CoT}
& MATH-Vision~\cite{wang2024measuring} \\
& We-Math~\cite{qiao2025we} \\
& LogicVista~\cite{xiao2024logicvista} \\
& DynaMath~\cite{zou2024dynamath} \\
& MMStar~\cite{chen2024we} \\
& MME~\cite{liang2024survey} \\
& MMBench-EN~\cite{liu2024mmbench} \\
& ScienceQA-IMG~\cite{saikh2022scienceqa} \\
& AI2D~\cite{kembhavi2016diagram} \\
& ChartQA~\cite{masry2022chartqa} \\
& InfoVQA~\cite{mathew2022infographicvqa} \\
& OCRBench~\cite{liu2024ocrbench} \\

\bottomrule
\end{tabular}

\caption{Evaluation benchmarks used for each dataset.}
\label{tab:benchmark_list}
\label{tab:eval_benchmarks}

\end{table*}

\begin{table*}[htbp] 
  \centering

  \begin{subtable}{\textwidth}
    \centering

    \resizebox{\textwidth}{!}{%
      \begin{tabular}{lcccccccccccccc}
      \toprule
      \textbf{Method} & \textbf{GQA} & \textbf{VizWiz} & \textbf{TextVQA} & \textbf{SQA-I} & \textbf{MME} & \textbf{MMB-CN} & \textbf{MMB-EN} & \textbf{AI2D} & \textbf{ChartQA} & \textbf{DocVQA} & \textbf{InfoVQA} & \textbf{MMStar} & \textbf{OCRBench} & \textbf{ARP} \\
      \midrule
        Full Data & 47.30 & 55.88 & 35.83 & 61.43 & 1270.40 & 50.26 & 55.67 & 52.14 & 15.68 & 15.80 & 15.53 & 35.32 & 26.20 & 100.00 \\
        \midrule
        Random & 42.93 & 55.79 & 36.54 & 60.34 & 1098.02 & 36.40 & 26.19 & 37.82 & 15.96 & 16.46 & 19.99 & 34.77 & 27.80 & 92.96 \\
        XMAS & 40.38 & 54.15 & 37.06 & 62.42 & 705.48 & 39.95 & 39.95 & 41.22 & 15.40 & 17.98 & 19.80 & 34.37 & 27.50 & 93.46 \\
        COINCIDE & 42.15 & 53.48 & 38.00 & 55.03 & 1000.71 & 29.12 & 41.24 & 36.40 & 16.64 & 17.89 & 17.79 & 33.51 & 28.70 & 92.26 \\
        D2Prune & 41.02 & \textbf{55.99} & 37.47 & 60.59 & 928.95 & 36.94 & 44.93 & 38.54 & 16.60 & 17.68 & 20.58 & 33.47 & 27.90 & 95.61 \\
        EL2N & 43.26 & 54.93 & 35.94 & 54.34 & 898.77 & 39.78 & 18.90 & 34.03 & 15.80 & 16.23 & 18.36 & 34.13 & 28.20 & 88.74 \\
        SemDeDup & 43.09 & 55.72 & 37.31 & \textbf{63.11} & 1115.77 & 41.32 & \textbf{48.71} & 41.16 & \textbf{17.80} & 19.02 & 19.42 & 33.09 & 27.80 & 99.45 \\
        CLIP & 39.24 & 54.77 & 31.80 & 57.71 & 549.88 & 38.06 & 40.38 & 37.63 & 15.92 & 16.78 & 19.69 & 33.75 & \textbf{29.20} & 89.92 \\
        PRISM & 29.97 & 52.06 & 26.73 & 49.33 & 1106.50 & 35.13 & 16.67 & 29.99 & 17.00 & 19.56 & 19.35 & \textbf{35.26} & 27.90 & 86.08 \\
        SCAL & 39.52 & 53.96 & 36.08 & 62.87 & 838.18 & 38.49 & 43.64 & \textbf{43.13} & 15.28 & 17.67 & 19.07 & 34.01 & 27.50 & 93.86 \\
        \rowcolor{lightcyan}
        MASS (Ours) & \textbf{43.61} & 55.45 & \textbf{38.92} & 59.84 & \textbf{1129.34} & \textbf{43.01} & 43.64 & 38.92 & 16.19 & \textbf{19.64} & \textbf{20.88} & 34.50 & 28.53 & \textbf{99.50} \\
      \bottomrule
      \end{tabular}%
    }
    \caption{5\% data subset.}
    \label{tab:vf_mass_subset_5}
  \end{subtable}


  \begin{subtable}{\textwidth}
    \centering

    \resizebox{\textwidth}{!}{%
      \begin{tabular}{lcccccccccccccc}
      \toprule
      \textbf{Method} & \textbf{GQA} & \textbf{VizWiz} & \textbf{TextVQA} & \textbf{SQA-I} & \textbf{MME} & \textbf{MMB-CN} & \textbf{MMB-EN} & \textbf{AI2D} & \textbf{ChartQA} & \textbf{DocVQA} & \textbf{InfoVQA} & \textbf{MMStar} & \textbf{OCRBench} & \textbf{ARP} \\
      \midrule
        Full Data & 47.30 & 55.88 & 35.83 & 61.43 & 1270.40 & 50.26 & 55.67 & 52.14 & 15.68 & 15.80 & 15.53 & 35.32 & 26.20 & 100.00 \\
        \midrule
        Random & 43.31 & 56.02 & 36.13 & 60.92 & 1152.77 & 39.61 & 31.88 & 44.50 & \textbf{16.52} & 17.27 & 18.08 & 34.36 & 27.90 & 95.30 \\
        XMAS & 42.73 & 55.82 & 36.55 & \textbf{64.40} & 1008.41 & 45.19 & 54.38 & 43.95 & 16.20 & 18.39 & 18.19 & 34.28 & 28.50 & 99.31 \\
        COINCIDE & 43.33 & \textbf{57.38} & 38.70 & 60.83 & 1032.00 & 43.56 & 41.24 & 40.16 & 15.84 & 18.96 & 19.25 & 34.27 & 29.00 & 97.93 \\
        D2Prune & 43.47 & 56.40 & 37.81 & 63.01 & 1146.50 & \textbf{47.68} & 51.80 & 43.01 & 16.44 & 17.38 & 18.78 & 34.73 & 29.00 & 100.49 \\
        EL2N & 44.08 & 56.99 & 38.10 & 47.99 & 1088.62 & 39.52 & 31.70 & 29.92 & 15.64 & 18.78 & 18.78 & 34.94 & 28.30 & 92.67 \\
        SemDeDup & 43.93 & 56.39 & 37.20 & 58.65 & \textbf{1236.25} & 47.51 & 45.96 & 39.31 & 15.72 & 18.89 & 17.58 & \textbf{36.00} & 28.50 & 98.97 \\
        CLIP & 41.02 & 53.63 & 38.11 & 59.84 & 725.09 & 34.28 & 30.50 & 35.56 & 16.36 & 20.29 & 22.72 & 34.90 & \textbf{29.10} & 94.13 \\
        PRISM & 35.21 & 48.80 & 37.87 & 61.97 & 751.61 & 38.49 & 44.16 & \textbf{44.88} & 16.44 & \textbf{23.76} & \textbf{23.37} & 35.92 & 28.80 & 98.99 \\
        SCAL & 42.73 & 56.26 & 37.68 & 60.88 & 1101.46 & 44.42 & \textbf{54.81} & 44.49 & 14.84 & 18.29 & 16.71 & 33.56 & 27.80 & 97.94 \\
        \rowcolor{lightcyan}
        MASS (Ours) & \textbf{45.05} & 56.05 & \textbf{38.83} & 63.25 & 1220.46 & 46.02 & 48.28 & 44.23 & 16.11 & 19.43 & 19.62 & 35.49 & 28.53 & \textbf{102.12} \\
      \bottomrule
      \end{tabular}%
    }
    \caption{10\% data subset.}
    \label{tab:vf_mass_subset_10}
  \end{subtable}


  \begin{subtable}{\textwidth}
    \centering

    \resizebox{\textwidth}{!}{%
      \begin{tabular}{lcccccccccccccc}
      \toprule
      \textbf{Method} & \textbf{GQA} & \textbf{VizWiz} & \textbf{TextVQA} & \textbf{SQA-I} & \textbf{MME} & \textbf{MMB-CN} & \textbf{MMB-EN} & \textbf{AI2D} & \textbf{ChartQA} & \textbf{DocVQA} & \textbf{InfoVQA} & \textbf{MMStar} & \textbf{OCRBench} & \textbf{ARP} \\
      \midrule
        Full Data & 47.30 & 55.88 & 35.83 & 61.43 & 1270.40 & 50.26 & 55.67 & 52.14 & 15.68 & 15.80 & 15.53 & 35.32 & 26.20 & 100.00 \\
        \midrule
        Random & 43.19 & 56.33 & 35.41 & 59.59 & 1230.04 & 42.44 & 38.06 & 41.06 & 15.52 & 16.26 & 16.26 & 32.27 & 27.60 & 93.82 \\
        XMAS & 44.68 & 56.52 & 37.19 & 61.18 & 1012.67 & 40.72 & 46.99 & 44.62 & 15.84 & 17.83 & 17.88 & 34.95 & 27.50 & 97.13 \\
        COINCIDE & 44.28 & 56.44 & 37.52 & 55.73 & 1221.20 & 45.96 & 39.26 & 37.50 & 17.24 & 19.03 & 18.32 & 35.59 & 28.80 & 98.39 \\
        D2Prune & 43.96 & 57.16 & 38.40 & 61.63 & 1059.64 & 44.67 & \textbf{49.48} & 44.66 & 15.80 & 17.62 & 18.60 & 34.13 & 28.30 & 98.94 \\
        EL2N & 45.81 & 57.45 & 38.52 & 53.69 & 1105.20 & 46.22 & 39.18 & 43.39 & 14.68 & 19.06 & 17.35 & 34.56 & 28.40 & 96.87 \\
        SemDeDup & \textbf{50.00} & 51.57 & 37.30 & 59.69 & \textbf{1247.46} & 43.90 & 45.45 & 43.13 & 17.04 & 18.66 & 18.93 & 35.24 & 28.00 & 100.35 \\
        CLIP & 41.33 & 55.10 & 38.44 & \textbf{65.25} & 640.09 & 39.60 & 41.67 & \textbf{48.41} & 16.68 & 20.19 & \textbf{21.96} & \textbf{37.57} & \textbf{29.30} & 99.24 \\
        PRISM & 39.25 & 52.24 & 38.62 & 61.23 & 770.61 & 41.67 & 41.07 & 43.78 & \textbf{17.96} & \textbf{21.12} & 21.00 & 36.22 & 29.10 & 98.64 \\
        SCAL & 43.15 & \textbf{57.90} & 38.47 & 57.56 & 1062.93 & 46.91 & 43.99 & 36.40 & 15.96 & 18.82 & 16.69 & 35.53 & 27.70 & 96.65 \\
        \rowcolor{lightcyan}
        MASS (Ours) & 45.75 & 56.54 & \textbf{38.74} & 63.33 & 1230.49 & \textbf{49.43} & 47.88 & 44.68 & 16.37 & 19.14 & 20.32 & 35.47 & 28.60 & \textbf{103.23} \\
      \bottomrule
      \end{tabular}%
    }
    \caption{15\% data subset.}
    \label{tab:vf_mass_subset_15}
  \end{subtable}
  \caption{Main experimental results on Vision-Flan using LLaVA-v1.5-7B as the target model. The best result in each column is highlighted in bold.}
  \label{tab:full_comparison_vf_mass}

\end{table*}

\begin{table*}[htbp]
  \centering

  \begin{subtable}{\textwidth}
    \centering

    \resizebox{\textwidth}{!}{%
      \begin{tabular}{lccccccccccccc}
      \toprule
      \textbf{Method} & \textbf{MATH-Vision} & \textbf{We-Math} & \textbf{LogicVista} & \textbf{DynaMath} & \textbf{MMStar} & \textbf{MME} & \textbf{MMBench-EN} & \textbf{ScienceQA-IMG} & \textbf{AI2D} & \textbf{ChartQA} & \textbf{InfoVQA} & \textbf{OCRBench} & \textbf{ARP} \\
      \midrule
        Full Data & 17.01 & 31.90 & 29.91 & 16.23 & 56.99 & 1549.25 & 77.92 & 95.44 & 76.94 & 85.68 & 62.29 & 73.50 & 100.00 \\
        \midrule
        Random & 14.08 & 27.99 & 26.12 & 16.07 & 52.35 & 1405.82 & 76.37 & 90.18 & 73.83 & \textbf{83.88} & 65.16 & 71.50 & 93.98 \\
        XMAS & 15.69 & 33.33 & 32.14 & \textbf{16.11} & 54.28 & 1301.91 & 76.80 & 90.38 & 73.93 & 83.68 & 64.51 & 71.20 & 97.51 \\
        COINCIDE & \textbf{17.63} & 38.22 & 31.03 & 15.27 & 52.86 & 1325.47 & 75.00 & 91.18 & \textbf{74.81} & 83.84 & 64.76 & 71.50 & 98.98 \\
        D2Prune & 17.40 & 34.14 & 28.35 & 14.37 & 52.22 & 1271.98 & 76.03 & 90.18 & 74.68 & 82.96 & 64.44 & 73.90 & 96.36 \\
        EL2N & 15.86 & 38.97 & \textbf{33.26} & 15.53 & 52.78 & 1378.95 & 75.95 & 91.03 & 74.19 & 83.40 & 65.34 & 69.20 & 99.13 \\
        SemDeDup & 15.33 & 38.91 & 28.35 & 15.49 & 53.56 & 1334.17 & 76.72 & 90.43 & 73.02 & 83.80 & 65.17 & 70.90 & 97.45 \\
        CLIP & 17.40 & 38.28 & 25.00 & 15.47 & 54.63 & 1258.75 & 76.80 & \textbf{91.42} & 72.73 & 80.72 & 62.10 & \textbf{75.50} & 96.98 \\
        PRISM & 17.01 & \textbf{40.63} & 29.02 & 13.43 & \textbf{54.79} & 1353.37 & 75.09 & 89.94 & 72.60 & 77.28 & 66.10 & 74.30 & 97.75 \\
        SCAL & 16.32 & 24.48 & 26.79 & 11.50 & 52.82 & 1445.83 & 76.55 & 89.04 & 72.93 & 67.08 & \textbf{67.69} & 74.90 & 91.19 \\
        \rowcolor{lightcyan}
        MASS (Ours) & 15.95 & 37.22 & 31.69 & 14.60 & 54.26 & \textbf{1470.73} & \textbf{77.61} & 90.73 & 74.35 & 83.79 & 67.60 & 74.70 & \textbf{99.65} \\
      \bottomrule
      \end{tabular}%
    }
    \caption{5\% data subset.}
    \label{tab:llavacot_llama32_11b_subset_5}
  \end{subtable}


  \begin{subtable}{\textwidth}
    \centering

    \resizebox{\textwidth}{!}{%
      \begin{tabular}{lccccccccccccc}
      \toprule
      \textbf{Method} & \textbf{MATH-Vision} & \textbf{We-Math} & \textbf{LogicVista} & \textbf{DynaMath} & \textbf{MMStar} & \textbf{MME} & \textbf{MMBench-EN} & \textbf{ScienceQA-IMG} & \textbf{AI2D} & \textbf{ChartQA} & \textbf{InfoVQA} & \textbf{OCRBench} & \textbf{ARP} \\
      \midrule
        Full Data & 17.01 & 31.90 & 29.91 & 16.23 & 56.99 & 1549.25 & 77.92 & 95.44 & 76.94 & 85.68 & 62.29 & 73.50 & 100.00 \\
        \midrule
        Random & 15.89 & 29.20 & 30.80 & 15.19 & 50.67 & 1471.44 & 76.55 & 91.77 & 74.48 & 84.12 & 66.30 & 74.20 & 96.85 \\
        XMAS & 16.97 & 35.63 & \textbf{32.59} & 15.73 & 53.52 & 1466.86 & 76.98 & 92.02 & 74.03 & 84.68 & 64.31 & 72.10 & 99.79 \\
        COINCIDE & 14.67 & 31.78 & 31.47 & \textbf{16.19} & 49.38 & 1465.80 & 77.58 & \textbf{92.07} & \textbf{76.04} & \textbf{84.80} & 65.57 & 74.20 & 97.68 \\
        D2Prune & 16.38 & 31.09 & 30.36 & 15.37 & 52.11 & 1449.64 & 77.23 & 91.08 & 74.81 & 83.76 & 64.68 & 72.20 & 97.21 \\
        EL2N & 16.05 & 35.06 & 31.47 & 15.57 & 52.72 & 1365.12 & 76.89 & 91.67 & 75.19 & 84.28 & 65.30 & 72.70 & 98.38 \\
        SemDeDup & 15.82 & 37.53 & 30.80 & 15.59 & 52.69 & \textbf{1528.53} & 76.37 & 91.57 & 75.00 & 84.44 & 64.18 & 74.00 & 99.54 \\
        CLIP & \textbf{17.83} & 30.17 & 28.57 & 15.25 & 53.22 & 1455.55 & 76.55 & 91.22 & 75.29 & 81.12 & 62.29 & 73.60 & 96.89 \\
        PRISM & 16.74 & \textbf{39.43} & 31.92 & 13.69 & 53.56 & 1426.71 & 76.12 & 90.18 & 73.12 & 81.04 & \textbf{68.66} & 74.70 & 99.40 \\
        SCAL & 13.91 & 32.47 & 26.79 & 13.43 & 50.21 & 1470.95 & 76.46 & 90.68 & 74.22 & 63.96 & 67.81 & \textbf{74.90} & 92.83 \\
        \rowcolor{lightcyan}
        MASS (Ours) & 16.72 & 33.83 & 31.55 & 15.32 & \textbf{54.85} & 1501.02 & \textbf{77.75} & 90.76 & 75.25 & 84.39 & 67.23 & 74.60 & \textbf{99.83} \\
      \bottomrule
      \end{tabular}%
    }
    \caption{10\% data subset.}
    \label{tab:llavacot_llama32_11b_subset_10}
  \end{subtable}


  \begin{subtable}{\textwidth}
    \centering

    \resizebox{\textwidth}{!}{%
      \begin{tabular}{lccccccccccccc}
      \toprule
      \textbf{Method} & \textbf{MATH-Vision} & \textbf{We-Math} & \textbf{LogicVista} & \textbf{DynaMath} & \textbf{MMStar} & \textbf{MME} & \textbf{MMBench-EN} & \textbf{ScienceQA-IMG} & \textbf{AI2D} & \textbf{ChartQA} & \textbf{InfoVQA} & \textbf{OCRBench} & \textbf{ARP} \\
      \midrule
        Full Data & 17.01 & 31.90 & 29.91 & 16.23 & 56.99 & 1549.25 & 77.92 & 95.44 & 76.94 & 85.68 & 62.29 & 73.50 & 100.00 \\
        \midrule
        Random & 15.56 & 29.54 & 32.37 & 15.02 & 53.44 & 1530.66 & 77.75 & 91.03 & 74.09 & \textbf{85.44} & 66.42 & 74.80 & 98.08 \\
        XMAS & 17.30 & 32.41 & 32.37 & 15.23 & 52.77 & 1485.83 & 76.80 & \textbf{92.27} & 74.61 & 84.68 & 65.48 & 74.20 & 99.25 \\
        COINCIDE & 17.04 & 37.99 & 31.25 & 15.35 & 51.78 & 1494.57 & 76.72 & 92.07 & 75.36 & 84.16 & 65.44 & 74.20 & 100.23 \\
        D2Prune & 16.74 & 27.30 & 33.71 & \textbf{15.99} & 53.32 & 1477.34 & 76.89 & 90.98 & 74.29 & 84.68 & 64.91 & 74.90 & 98.30 \\
        EL2N & 17.14 & \textbf{43.56} & 31.47 & 15.29 & 53.19 & \textbf{1570.67} & 77.75 & 91.03 & \textbf{75.74} & 84.48 & 65.95 & 73.50 & 102.46 \\
        SemDeDup & 16.78 & 34.89 & 30.58 & 15.79 & 53.32 & 1478.05 & 77.23 & 91.52 & 75.06 & 84.32 & 64.90 & 74.40 & 99.41 \\
        CLIP & \textbf{18.19} & 39.08 & 33.71 & 14.43 & \textbf{55.46} & 1477.28 & 76.55 & 91.47 & 73.45 & 82.12 & 64.22 & \textbf{75.70} & 101.27 \\
        PRISM & 16.18 & 37.01 & 31.70 & 12.10 & 53.14 & 1494.04 & 77.15 & 90.83 & 74.42 & 83.12 & 67.87 & 73.80 & 98.22 \\
        SCAL & 13.39 & 25.98 & 31.47 & 15.47 & 51.14 & 1505.28 & \textbf{78.09} & 91.77 & 74.61 & 68.56 & \textbf{68.94} & 74.00 & 94.36 \\
        \rowcolor{lightcyan}
        MASS (Ours) & 16.83 & 39.75 & \textbf{36.09} & 15.60 & 53.55 & 1507.36 & \textbf{78.09} & 91.69 & 75.04 & 84.92 & 67.09 & 74.67 & \textbf{102.82} \\
      \bottomrule
      \end{tabular}%
    }
    \caption{15\% data subset.}
    \label{tab:llavacot_llama32_11b_subset_15}
  \end{subtable}
  \caption{Main experimental results on LLaVA-CoT using Llama-3.2-11B-Vision-Instruct as the target model. The best result in each column is highlighted in bold.}
  \label{tab:full_comparison_llavacot_mass}
\end{table*}

\begin{table*}[htbp]
  \centering

  \begin{subtable}{\textwidth}
    \centering
    \label{tab:dae_manifold_ablation_vf_subset_10}
    \resizebox{\textwidth}{!}{%
      \begin{tabular}{lcccccccccccccc}
      \toprule
      \textbf{Method} & \textbf{GQA} & \textbf{VizWiz} & \textbf{TextVQA} & \textbf{SQA-I} & \textbf{MME} & \textbf{MMB-CN} & \textbf{MMB-EN} & \textbf{AI2D} & \textbf{ChartQA} & \textbf{DocVQA} & \textbf{InfoVQA} & \textbf{MMStar} & \textbf{OCRBench} & \textbf{ARP} \\
      \midrule
        Full Data & 47.30 & 55.88 & 35.83 & 61.43 & 1270.40 & 50.26 & 55.67 & 52.14 & 15.68 & 15.80 & 15.53 & 35.32 & 26.20 & 100.00 \\
        \midrule
        RAW  & \textbf{45.33} & 53.91 & 36.41 & 62.87 & \textbf{1291.65} & 44.33 & \textbf{51.03} & \textbf{46.11} & \textbf{16.40} & 17.86 & \textbf{20.07} & 35.18 & 28.30 & 101.59 \\
        DAE  & 45.05 & \textbf{56.05} & \textbf{38.83} & \textbf{63.25} & 1220.46 & \textbf{46.02} & 48.28 & 44.23 & 16.11 & \textbf{19.43} & 19.62 & \textbf{35.49} & \textbf{28.53} & \textbf{102.12} \\
      \bottomrule
      \end{tabular}%
    }
  \end{subtable}
  \caption{Ablation results of DAE manifold coordinates. We compare MASS with raw embedding clustering and DAE principal manifold clustering on the Vision-Flan dataset using LLaVA-v1.5-7B with a 10\% data subset. The best result in each column is highlighted in bold.}
  \label{tab:dae_manifold_ablation_vf}

\end{table*}

\begin{table*}[htbp]
  \centering

  \begin{subtable}{\textwidth}
    \centering
    \resizebox{\textwidth}{!}{%
      \begin{tabular}{lcccccccccccccc}
      \toprule
      \textbf{Method} & \textbf{GQA} & \textbf{VizWiz} & \textbf{TextVQA} & \textbf{SQA-I} & \textbf{MME} & \textbf{MMB-CN} & \textbf{MMB-EN} & \textbf{AI2D} & \textbf{ChartQA} & \textbf{DocVQA} & \textbf{InfoVQA} & \textbf{MMStar} & \textbf{OCRBench} & \textbf{ARP} \\
      \midrule
        Full Data & 47.30 & 55.88 & 35.83 & 61.43 & 1270.40 & 50.26 & 55.67 & 52.14 & 15.68 & 15.80 & 15.53 & 35.32 & 26.20 & 100.00 \\
        \midrule
        Random & 43.31 & 56.02 & 36.13 & 60.92 & 1152.77 & 39.61 & 31.88 & \textbf{44.50} & 16.52 & 17.27 & 18.08 & 34.36 & 27.90 & 95.30 \\
        w/o SAE & 43.85 & 55.30 & 37.93 & 59.10 & \textbf{1236.01} & 42.61 & 40.72 & 37.92 & \textbf{17.00} & 18.73 & 17.70 & 35.19 & 28.30 & 97.71 \\
        w/o DAE & 44.47 & 53.99 & \textbf{40.14} & 57.21 & 1184.86 & 44.85 & 40.21 & 41.90 & 16.20 & \textbf{22.07} & \textbf{21.33} & 34.05 & \textbf{29.10} & 101.44 \\
        ALL & \textbf{45.05} & \textbf{56.05} & 38.83 & \textbf{63.25} & 1220.46 & \textbf{46.02} & \textbf{48.28} & 44.23 & 16.11 & 19.43 & 19.62 & \textbf{35.49} & 28.53 & \textbf{102.12} \\
      \bottomrule
      \end{tabular}%
    }
  \end{subtable}
  \caption{Ablation results of the complementarity of DAE and SAE. We evaluate random selection, DAE-only grouping, SAE-only coverage selection, and the full MASS method on the Vision-Flan dataset using LLaVA-v1.5-7B with a 10\% data subset. The best result in each column is highlighted in bold.}
 \label{tab:dae_sae_complementarity_vf}

\end{table*}

\clearpage

\begin{table*}[htbp]
  \centering

  \begin{subtable}{\textwidth}
    \centering
    \label{tab:dae_latent_dim_sensitivity_vf_subset_10}
    \resizebox{\textwidth}{!}{%
      \begin{tabular}{lcccccccccccccc}
      \toprule
      \textbf{Method} & \textbf{GQA} & \textbf{VizWiz} & \textbf{TextVQA} & \textbf{SQA-I} & \textbf{MME} & \textbf{MMB-CN} & \textbf{MMB-EN} & \textbf{AI2D} & \textbf{ChartQA} & \textbf{DocVQA} & \textbf{InfoVQA} & \textbf{MMStar} & \textbf{OCRBench} & \textbf{ARP} \\
      \midrule
        Full Data & 47.30 & 55.88 & 35.83 & 61.43 & 1270.40 & 50.26 & 55.67 & 52.14 & 15.68 & 15.80 & 15.53 & 35.32 & 26.20 & 100.00 \\
        \midrule
        Random & 43.31 & 56.02 & 36.13 & 60.92 & 1152.77 & 39.61 & 31.88 & 44.50 & 16.52 & 17.27 & 18.08 & 34.36 & 27.90 & 95.30 \\
        8 & 44.47 & \textbf{57.80} & 38.81 & 60.78 & 1182.15 & \textbf{48.11} & 46.22 & 42.94 & 15.80 & 19.25 & 18.80 & 34.91 & 28.10 & 100.66 \\
        16 & 44.49 & 56.08 & 37.66 & 60.83 & 1226.66 & 45.36 & 43.21 & 40.67 & 15.40 & 18.89 & 18.87 & \textbf{35.56} & 27.30 & 98.86 \\
        32 & 45.05 & 56.05 & 38.83 & 63.25 & 1220.46 & 46.02 & 48.28 & 44.23 & 16.11 & 19.43 & 19.62 & 35.49 & \textbf{28.53} & 102.12 \\
        64 & \textbf{45.68} & 56.93 & \textbf{39.55} & \textbf{63.61} & 1149.54 & 46.74 & \textbf{51.37} & \textbf{44.92} & 16.00 & 19.33 & \textbf{20.26} & 35.19 & 27.90 & \textbf{102.71} \\
        128 & 44.95 & 55.84 & 38.19 & 60.39 & \textbf{1231.39} & 45.67 & 47.74 & 42.04 & \textbf{16.64} & \textbf{19.81} & 20.02 & 35.32 & 28.43 & 101.77 \\
      \bottomrule
      \end{tabular}%
    }
  \end{subtable}
  \caption{Sensitivity analysis results of DAE latent dimension. We vary the DAE latent dimension on the Vision-Flan dataset using LLaVA-v1.5-7B with a 10\% data subset, while keeping all other MASS settings unchanged. The best result in each column is highlighted in bold.}
  \label{tab:dae_latent_dim_sensitivity_vf}

\end{table*}

\begin{table*}[htbp]
  \centering


  \begin{subtable}{\textwidth}
    \centering

    \resizebox{\textwidth}{!}{%
      \begin{tabular}{lcccccccccccccc}
      \toprule
      \textbf{Method} & \textbf{GQA} & \textbf{VizWiz} & \textbf{TextVQA} & \textbf{SQA-I} & \textbf{MME} & \textbf{MMB-CN} & \textbf{MMB-EN} & \textbf{AI2D} & \textbf{ChartQA} & \textbf{DocVQA} & \textbf{InfoVQA} & \textbf{MMStar} & \textbf{OCRBench} & \textbf{ARP} \\
      \midrule
        Full Data & 47.30 & 55.88 & 35.83 & 61.43 & 1270.40 & 50.26 & 55.67 & 52.14 & 15.68 & 15.80 & 15.53 & 35.32 & 26.20 & 100.00 \\
        \midrule
        65536 & \textbf{43.63} & 55.32 & 37.91 & \textbf{62.17} & \textbf{1178.10} & 39.89 & \textbf{44.96} & \textbf{41.63} & 15.92 & 18.78 & 20.38 & \textbf{35.12} & 28.17 & 99.19 \\
        131072 & 43.61 & \textbf{55.45} & \textbf{38.92} & 59.84 & 1129.34 & \textbf{43.01} & 43.64 & 38.92 & \textbf{16.19} & \textbf{19.64} & \textbf{20.88} & 34.50 & \textbf{28.53} & \textbf{99.50} \\
      \bottomrule
      \end{tabular}%
    }
    \caption{Effect of SAE feature dimension under the 5\% data subset.}
    \label{tab:sae_dim_sensitivity_vf_subset_5}
  \end{subtable}


  \begin{subtable}{\textwidth}
    \centering

    \resizebox{\textwidth}{!}{%
      \begin{tabular}{lcccccccccccccc}
      \toprule
      \textbf{Method} & \textbf{GQA} & \textbf{VizWiz} & \textbf{TextVQA} & \textbf{SQA-I} & \textbf{MME} & \textbf{MMB-CN} & \textbf{MMB-EN} & \textbf{AI2D} & \textbf{ChartQA} & \textbf{DocVQA} & \textbf{InfoVQA} & \textbf{MMStar} & \textbf{OCRBench} & \textbf{ARP} \\
      \midrule
        Full Data & 47.30 & 55.88 & 35.83 & 61.43 & 1270.40 & 50.26 & 55.67 & 52.14 & 15.68 & 15.80 & 15.53 & 35.32 & 26.20 & 100.00 \\
        \midrule
        65536 & 44.42 & \textbf{56.17} & 37.64 & \textbf{63.54} & 1166.15 & 45.10 & \textbf{49.86} & 43.09 & \textbf{16.17} & 18.80 & 19.56 & 35.35 & 28.27 & 100.98 \\
        131072 & \textbf{45.05} & 56.05 & \textbf{38.83} & 63.25 & \textbf{1220.46} & \textbf{46.02} & 48.28 & \textbf{44.23} & 16.11 & \textbf{19.43} & \textbf{19.62} & \textbf{35.49} & \textbf{28.53} & \textbf{102.12} \\
      \bottomrule
      \end{tabular}%
    }
    \caption{Effect of SAE feature dimension under the 10\% data subset.}
    \label{tab:sae_dim_sensitivity_vf_subset_10}
  \end{subtable}


  \begin{subtable}{\textwidth}
    \centering

    \resizebox{\textwidth}{!}{%
      \begin{tabular}{lcccccccccccccc}
      \toprule
      \textbf{Method} & \textbf{GQA} & \textbf{VizWiz} & \textbf{TextVQA} & \textbf{SQA-I} & \textbf{MME} & \textbf{MMB-CN} & \textbf{MMB-EN} & \textbf{AI2D} & \textbf{ChartQA} & \textbf{DocVQA} & \textbf{InfoVQA} & \textbf{MMStar} & \textbf{OCRBench} & \textbf{ARP} \\
      \midrule
        Full Data & 47.30 & 55.88 & 35.83 & 61.43 & 1270.40 & 50.26 & 55.67 & 52.14 & 15.68 & 15.80 & 15.53 & 35.32 & 26.20 & 100.00 \\
        \midrule
        65536 & 44.52 & \textbf{57.04} & 38.52 & 61.38 & \textbf{1251.89} & 48.82 & \textbf{48.77} & \textbf{45.41} & 16.28 & 19.08 & 19.24 & \textbf{36.60} & \textbf{28.93} & 102.81 \\
        131072 & \textbf{45.75} & 56.54 & \textbf{38.74} & \textbf{63.33} & 1230.49 & \textbf{49.43} & 47.88 & 44.68 & \textbf{16.37} & \textbf{19.14} & \textbf{20.32} & 35.47 & 28.60 & \textbf{103.23} \\
      \bottomrule
      \end{tabular}%
    }
    \caption{Effect of SAE feature dimension under the 15\% data subset.}
    \label{tab:sae_dim_sensitivity_vf_subset_15}
  \end{subtable}



  \begin{subtable}{\textwidth}
    \centering

    \resizebox{\textwidth}{!}{%
      \begin{tabular}{lcccccccccccccc}
      \toprule
      \textbf{Method} & \textbf{GQA} & \textbf{VizWiz} & \textbf{TextVQA} & \textbf{SQA-I} & \textbf{MME} & \textbf{MMB-CN} & \textbf{MMB-EN} & \textbf{AI2D} & \textbf{ChartQA} & \textbf{DocVQA} & \textbf{InfoVQA} & \textbf{MMStar} & \textbf{OCRBench} & \textbf{ARP} \\
      \midrule
        Full Data & 47.30 & 55.88 & 35.83 & 61.43 & 1270.40 & 50.26 & 55.67 & 52.14 & 15.68 & 15.80 & 15.53 & 35.32 & 26.20 & 100.00 \\
        \midrule
        16 & 44.58 & \textbf{57.19} & 38.73 & 61.36 & 1226.62 & 44.19 & 45.31 & 41.23 & 15.96 & \textbf{19.80} & 19.88 & 35.25 & 27.80 & 100.81 \\
        32 & 44.59 & 56.41 & 37.54 & 60.77 & \textbf{1261.05} & \textbf{47.22} & 48.20 & 41.22 & 15.91 & 17.87 & 18.87 & 34.35 & \textbf{28.90} & 100.11 \\
        64 & \textbf{45.05} & 56.05 & \textbf{38.83} & \textbf{63.25} & 1220.46 & 46.02 & \textbf{48.28} & \textbf{44.23} & \textbf{16.11} & 19.43 & 19.62 & \textbf{35.49} & 28.53 & \textbf{102.12} \\
        128 & 44.96 & 56.90 & 38.61 & 61.84 & 1168.17 & 45.39 & 46.82 & 42.31 & \textbf{16.11} & 19.24 & \textbf{20.40} & 34.60 & 28.57 & 101.21 \\
      \bottomrule
      \end{tabular}%
    }
    \caption{Effect of TopK sparsity under the 10\% data subset with SAE feature dimension fixed to 128.}
    \label{tab:sae_topk_sensitivity_vf_subset_10}
  \end{subtable}
  \caption{Sensitivity analysis results of SAE configuration. We analyze the effect of SAE feature dimension under 5\%, 10\%, and 15\% data budgets, and the effect of TopK sparsity under the 10\% data budget on the Vision-Flan dataset using LLaVA-v1.5-7B. The best result in each column is highlighted in bold.}
  \label{tab:sae_configuration_sensitivity_vf}
\end{table*}

\begin{table*}[htbp]
  \centering

  \begin{subtable}{\textwidth}
    \centering
    \label{tab:quality_signal_ablation_vf_subset_10}
    \resizebox{\textwidth}{!}{%
      \begin{tabular}{lcccccccccccccc}
      \toprule
      \textbf{Method} & \textbf{GQA} & \textbf{VizWiz} & \textbf{TextVQA} & \textbf{SQA-I} & \textbf{MME} & \textbf{MMB-CN} & \textbf{MMB-EN} & \textbf{AI2D} & \textbf{ChartQA} & \textbf{DocVQA} & \textbf{InfoVQA} & \textbf{MMStar} & \textbf{OCRBench} & \textbf{ARP} \\
      \midrule
        Full Data & 47.30 & 55.88 & 35.83 & 61.43 & 1270.40 & 50.26 & 55.67 & 52.14 & 15.68 & 15.80 & 15.53 & 35.32 & 26.20 & 100.00 \\
        \midrule
        Random & 43.31 & 56.02 & 36.13 & 60.92 & 1152.77 & 39.61 & 31.88 & 44.50 & \textbf{16.52} & 17.27 & 18.08 & 34.36 & 27.90 & 95.30 \\
        w/o Quality & 43.29 & 55.84 & 38.44 & 62.96 & 1048.93 & \textbf{46.13} & 50.43 & 41.55 & 15.20 & \textbf{19.44} & \textbf{19.82} & 34.23 & \textbf{28.60} & 99.96 \\
        LLaVA-OneVision & \textbf{45.21} & 56.02 & \textbf{39.07} & \textbf{63.86} & 1202.30 & 45.53 & \textbf{52.58} & \textbf{45.01} & 15.96 & 19.24 & 19.29 & \textbf{35.72} & 28.30 & \textbf{102.44} \\
        Default & 45.05 & \textbf{56.05} & 38.83 & 63.25 & \textbf{1220.46} & 46.02 & 48.28 & 44.23 & 16.11 & 19.43 & 19.62 & 35.49 & 28.53 & 102.12 \\
      \bottomrule
      \end{tabular}%
    }
  \end{subtable}
  \caption{Ablation results of quality signals. We evaluate MASS without quality signals and with different external scoring models on the Vision-Flan dataset using LLaVA-v1.5-7B with a 10\% data subset. The best result in each column is highlighted in bold.}
  \label{tab:quality_signal_ablation_vf}
\end{table*}

\begin{table*}[htbp]
  \centering

  \begin{subtable}{\textwidth}
    \centering
    \label{tab:embedding_source_ablation_vf_subset_10}
    \resizebox{\textwidth}{!}{%
      \begin{tabular}{lcccccccccccccc}
      \toprule
      \textbf{Method} & \textbf{GQA} & \textbf{VizWiz} & \textbf{TextVQA} & \textbf{SQA-I} & \textbf{MME} & \textbf{MMB-CN} & \textbf{MMB-EN} & \textbf{AI2D} & \textbf{ChartQA} & \textbf{DocVQA} & \textbf{InfoVQA} & \textbf{MMStar} & \textbf{OCRBench} & \textbf{ARP} \\
      \midrule
        Full Data & 47.30 & 55.88 & 35.83 & 61.43 & 1270.40 & 50.26 & 55.67 & 52.14 & 15.68 & 15.80 & 15.53 & 35.32 & 26.20 & 100.00 \\
        \midrule
        Random & 43.31 & 56.02 & 36.13 & 60.92 & 1152.77 & 39.61 & 31.88 & \textbf{44.50} & \textbf{16.52} & 17.27 & 18.08 & 34.36 & \textbf{27.90} & 95.30 \\
        GME & \textbf{45.36} & \textbf{57.13} & \textbf{39.10} & \textbf{62.87} & \textbf{1240.33} & \textbf{45.96} & \textbf{51.46} & 40.32 & 16.04 & \textbf{19.32} & \textbf{19.47} & \textbf{34.91} & 27.40 & \textbf{101.68} \\
      \bottomrule
      \end{tabular}%
    }
  \end{subtable}
  \caption{Ablation results of the embedding source. We replace the original embedding source with GME on the Vision-Flan dataset using LLaVA-v1.5-7B with a 10\% data subset, while keeping all other MASS settings unchanged. The best result in each column is highlighted in bold.}
  \label{tab:embedding_source_ablation_vf}
\end{table*}

\begin{table*}[htbp]
  \centering

  \begin{subtable}{\textwidth}
    \centering

    \resizebox{\textwidth}{!}{%
      \begin{tabular}{lcccccccccccccc}
      \toprule
      \textbf{Method} & \textbf{GQA} & \textbf{VizWiz} & \textbf{TextVQA} & \textbf{SQA-I} & \textbf{MME} & \textbf{MMB-CN} & \textbf{MMB-EN} & \textbf{AI2D} & \textbf{ChartQA} & \textbf{DocVQA} & \textbf{InfoVQA} & \textbf{MMStar} & \textbf{OCRBench} & \textbf{ARP} \\
      \midrule
        Full Data & 49.02 & 55.92 & 35.82 & 70.60 & 1318.89 & 56.44 & 60.74 & 57.67 & 19.44 & 20.03 & 19.87 & 35.65 & 29.50 & 100.00 \\
        \midrule
        Random & 43.02 & 56.82 & 34.35 & 63.46 & 1178.91 & 36.25 & 41.23 & \textbf{45.66} & 14.88 & 14.87 & 18.17 & \textbf{32.17} & 29.00 & 85.12 \\
        MASS & \textbf{43.55} & \textbf{57.10} & \textbf{36.34} & \textbf{65.10} & \textbf{1222.08} & \textbf{41.67} & \textbf{43.13} & 42.39 & \textbf{16.64} & \textbf{17.92} & \textbf{20.79} & 31.75 & \textbf{29.20} & \textbf{89.49} \\
      \bottomrule
      \end{tabular}%
    }
    \caption{5\% data subset.}
    \label{tab:robustness_vf_llava15_13b_subset_5}
  \end{subtable}


  \begin{subtable}{\textwidth}
    \centering

    \resizebox{\textwidth}{!}{%
      \begin{tabular}{lcccccccccccccc}
      \toprule
      \textbf{Method} & \textbf{GQA} & \textbf{VizWiz} & \textbf{TextVQA} & \textbf{SQA-I} & \textbf{MME} & \textbf{MMB-CN} & \textbf{MMB-EN} & \textbf{AI2D} & \textbf{ChartQA} & \textbf{DocVQA} & \textbf{InfoVQA} & \textbf{MMStar} & \textbf{OCRBench} & \textbf{ARP} \\
      \midrule
        Full Data & 49.02 & 55.92 & 35.82 & 70.60 & 1318.89 & 56.44 & 60.74 & 57.67 & 19.44 & 20.03 & 19.87 & 35.65 & 29.50 & 100.00 \\
        \midrule
        Random & \textbf{44.98} & 58.13 & 36.43 & \textbf{68.26} & 1176.93 & \textbf{47.60} & \textbf{55.56} & 49.63 & 16.56 & 16.97 & 19.08 & 31.35 & 29.60 & 92.26 \\
        MASS & 44.85 & \textbf{58.45} & \textbf{36.45} & 68.17 & \textbf{1255.35} & 44.24 & 50.52 & \textbf{50.91} & \textbf{16.60} & \textbf{17.77} & \textbf{20.17} & \textbf{33.20} & \textbf{30.10} & \textbf{93.09} \\
      \bottomrule
      \end{tabular}%
    }
    \caption{10\% data subset.}
    \label{tab:robustness_vf_llava15_13b_subset_10}
  \end{subtable}


  \begin{subtable}{\textwidth}
    \centering

    \resizebox{\textwidth}{!}{%
      \begin{tabular}{lcccccccccccccc}
      \toprule
      \textbf{Method} & \textbf{GQA} & \textbf{VizWiz} & \textbf{TextVQA} & \textbf{SQA-I} & \textbf{MME} & \textbf{MMB-CN} & \textbf{MMB-EN} & \textbf{AI2D} & \textbf{ChartQA} & \textbf{DocVQA} & \textbf{InfoVQA} & \textbf{MMStar} & \textbf{OCRBench} & \textbf{ARP} \\
      \midrule
        Full Data & 49.02 & 55.92 & 35.82 & 70.60 & 1318.89 & 56.44 & 60.74 & 57.67 & 19.44 & 20.03 & 19.87 & 35.65 & 29.50 & 100.00 \\
        \midrule
        Random & 45.25 & \textbf{56.96} & \textbf{38.03} & 65.05 & \textbf{1176.15} & 42.70 & 46.99 & 48.32 & 16.60 & 18.38 & 18.23 & \textbf{34.86} & 29.40 & 91.14 \\
        MASS & \textbf{46.47} & 55.04 & 37.78 & \textbf{68.07} & 1150.51 & \textbf{46.82} & \textbf{49.31} & \textbf{51.65} & \textbf{17.16} & \textbf{18.60} & \textbf{22.71} & 31.86 & \textbf{30.40} & \textbf{94.15} \\
      \bottomrule
      \end{tabular}%
    }
    \caption{15\% data subset.}
    \label{tab:robustness_vf_llava15_13b_subset_15}
  \end{subtable}
  \caption{Robustness results across target models on Vision-Flan using LLaVA-v1.5-13B as the target model. The best result in each column is highlighted in bold.}
  \label{tab:robustness_vf_llava15_13b}
\end{table*}

\begin{table*}[htbp]
  \centering

  \begin{subtable}{\textwidth}
    \centering

    \resizebox{\textwidth}{!}{%
      \begin{tabular}{lccccccccc}
      \toprule
      \textbf{Method} & \textbf{MATH-Vision} & \textbf{MME} & \textbf{MMBench-EN} & \textbf{ScienceQA-IMG} & \textbf{AI2D} & \textbf{ChartQA} & \textbf{InfoVQA} & \textbf{OCRBench} & \textbf{ARP} \\
      \midrule
        Full Data & 9.05 & 1350.72 & 64.86 & 77.89 & 63.18 & 73.44 & 49.67 & 71.00 & 100.00 \\
        \midrule
        Random & 9.74 & 1233.91 & 58.08 & 71.19 & 59.29 & \textbf{69.00} & 45.24 & 67.90 & 94.30 \\
        MASS & \textbf{9.84} & \textbf{1277.98} & \textbf{62.20} & \textbf{72.53} & \textbf{60.59} & 67.52 & \textbf{46.60} & \textbf{69.70} & \textbf{96.52} \\
      \bottomrule
      \end{tabular}%
    }
    \caption{5\% data subset.}
    \label{tab:robustness_llavacot_qwen2vl_2b_subset_5}
  \end{subtable}


  \begin{subtable}{\textwidth}
    \centering

    \resizebox{\textwidth}{!}{%
      \begin{tabular}{lccccccccc}
      \toprule
      \textbf{Method} & \textbf{MATH-Vision} & \textbf{MME} & \textbf{MMBench-EN} & \textbf{ScienceQA-IMG} & \textbf{AI2D} & \textbf{ChartQA} & \textbf{InfoVQA} & \textbf{OCRBench} & \textbf{ARP} \\
      \midrule
        Full Data & 9.05 & 1350.72 & 64.86 & 77.89 & 63.18 & 73.44 & 49.67 & 71.00 & 100.00 \\
        \midrule
        Random & \textbf{9.80} & \textbf{1216.21} & 61.08 & 69.26 & 59.52 & \textbf{71.04} & 45.98 & 68.90 & 95.25 \\
        MASS & 9.31 & 1205.24 & \textbf{63.23} & \textbf{73.23} & \textbf{59.97} & 70.84 & \textbf{46.75} & \textbf{70.30} & \textbf{96.02} \\
      \bottomrule
      \end{tabular}%
    }
    \caption{10\% data subset.}
    \label{tab:robustness_llavacot_qwen2vl_2b_subset_10}
  \end{subtable}


  \begin{subtable}{\textwidth}
    \centering

    \resizebox{\textwidth}{!}{%
      \begin{tabular}{lccccccccc}
      \toprule
      \textbf{Method} & \textbf{MATH-Vision} & \textbf{MME} & \textbf{MMBench-EN} & \textbf{ScienceQA-IMG} & \textbf{AI2D} & \textbf{ChartQA} & \textbf{InfoVQA} & \textbf{OCRBench} & \textbf{ARP} \\
      \midrule
        Full Data & 9.05 & 1350.72 & 64.86 & 77.89 & 63.18 & 73.44 & 49.67 & 71.00 & 100.00 \\
        \midrule
        Random & \textbf{8.68} & \textbf{1282.13} & 59.97 & 71.99 & 59.49 & \textbf{71.52} & 46.41 & 69.20 & 94.77 \\
        MASS & 8.22 & 1268.12 & \textbf{62.63} & \textbf{73.38} & \textbf{60.36} & 70.36 & \textbf{48.49} & \textbf{69.90} & \textbf{95.36} \\
      \bottomrule
      \end{tabular}%
    }
    \caption{15\% data subset.}
    \label{tab:robustness_llavacot_qwen2vl_2b_subset_15}
  \end{subtable}
  \caption{Robustness results across target models on LLaVA-CoT using Qwen2VL-2B-Instruct as the target model. The best result in each column is highlighted in bold.}
  \label{tab:robustness_llavacot_qwen2vl_2b}
\end{table*}

\clearpage
\begin{table*}[t]
\centering

\begin{tabular}{lc}
\toprule
Method & GPU Hours \\
\midrule
XMAS & 15.99 \\
ScalSelect & 8.82 \\
PRISM & 6.70 \\
COINCIDE & 9.06 \\
D2 Prune & 12.51 \\
EL2N & 9.57 \\
SemDeDup & 9.55 \\
MASS (Ours) & 15.39 \\
\bottomrule
\end{tabular}
\caption{Time cost comparison of different methods on LLaVA-CoT.}
\label{tab:time_cost}
\end{table*}

\bibliography{aaai2027}


\end{document}